\documentclass[11pt]{article}
\usepackage[preprint]{acl}
\usepackage{times}
\usepackage{latexsym}
\usepackage[T1]{fontenc}
\usepackage[utf8]{inputenc}
\usepackage{microtype}
\usepackage{graphicx}
\usepackage{booktabs}
\usepackage{amsmath}
\usepackage{amssymb}
\usepackage{algorithm}
\usepackage{algpseudocode}
\newcommand{\fit}[1]{\resizebox{\ifdim\width>\columnwidth \columnwidth\else\width\fi}{!}{#1}}

\title{A Learning-Rate-Gated Failure of GRPO in a Small Language and Vision-Language Model Web Agent: A Controlled Null and Its Mechanism}

\author{
  \textbf{Chengguang Gan\textsuperscript{1}},
  \textbf{Zhixi Cai\textsuperscript{2}},
  \textbf{Yunhao Liang\textsuperscript{3}},
  \textbf{Hanjun Wei\textsuperscript{3}},
  \textbf{Shiwen Ni\textsuperscript{4}},
  \textbf{Qinghao Zhang\textsuperscript{5}}
\\
  \textsuperscript{1}Independent Researcher \quad
  \textsuperscript{2}Monash University \quad
  \textsuperscript{3}University of Chinese Academy of Sciences
\\
  \textsuperscript{4}Shenzhen University of Advanced Technology \quad
  \textsuperscript{5}Pusan National University
\\
  \small{\textbf{Correspondence:} \href{mailto:chengguangg1024@gmail.com}{chengguangg1024@gmail.com}}
}

\begin{document}
\raggedbottom
\maketitle
\begin{abstract}
Reinforcement learning with verifiable rewards, and Group Relative Policy Optimization (GRPO) in particular, is now run routinely on a supervised checkpoint in the hope of producing a stronger agent. We ask whether it adds skill to a small language and vision-language model web agent at the 4B to 8B scale, or whether it mostly reshapes behavior the supervised model already has. Across a control grid of 18 runs that varies learning rate, KL weight, seed, initialization, and clipping, no configuration credibly improves the success rate of a strong supervised baseline on tasks the agent has largely mastered. On the text track, moderate to high learning rates make it credibly worse. The null holds under paired testing, 25 evaluation seeds, 6 training seeds, changes to the recipe, both text and Set-of-Marks screenshot observations, and scaling the backbone to 8B; the credible harm is a text-track finding and is only nominal under Set-of-Marks. To show that the null reflects the setting and not a broken pipeline, we run the identical harness, reward, and recipe on tasks whose reward is reachable by sampling, and there the success rate rises by 22 points with a paired interval that excludes zero. GRPO therefore helps only when there is headroom to climb, meaning the sampled policy already succeeds more often than the greedy one. We then explain the failure. A middle learning rate degrades the agent and a high one collapses it, and the two regimes form a double dissociation: grafting localizes the degrade regime to the attention and MLP blocks, while the collapse regime cannot be traced to any single group, and the embedding change that dominates the weight movement is causally inert. At 4B, effective rank in the late layers tracks capability in both directions; at 8B the two come apart. This coupling is specific to the smaller model, so we report it as scale-dependent.
\end{abstract}

\section{Introduction}

\begin{figure}[!t]
\centering
\includegraphics[width=\columnwidth]{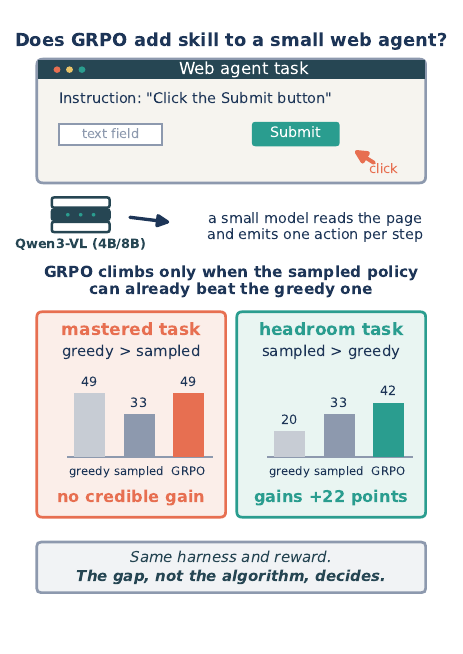}
\caption{The question and the answer at a glance. GRPO adds no credible gain on tasks the agent has already mastered, but the same recipe gains 22 points where the reward is reachable by sampling.}
\label{fig:teaser}
\end{figure}

Post-training with reinforcement learning has become routine for language and multimodal agents. A supervised model is first taught to follow the task format, and GRPO or a close relative \citep{shao2024deepseekmath,yu2026dapo,liu2025understanding} is then run against a verifiable reward in the hope of pushing capability past what supervision alone reached. For web agents the appeal is direct, since rollouts are cheap to score against task success and the recipe that lifted mathematics and coding models is expected to lift an agent that clicks, types, and navigates \citep{qi2025webrl}. What remains unclear is whether this second stage adds new skill to a small language and vision-language model agent, or whether it mostly sharpens choices the supervised model could already make \citep{chen2026does}. This matters in practice. If reinforcement learning adds capability, more of it is the right investment; if it only sharpens what the supervised model already does, the effort belongs in better supervision or distillation.

We study one such agent under tight control: a 4B language and vision-language model \citep{bai2025qwen3} driving a web agent on the MiniWoB benchmark \citep{shi2017world,liu2018reinforcement}. Holding the harness, reward, and task set fixed, we sweep the reinforcement-learning choices a practitioner would tune and score every run with clean greedy decoding and a paired test against the supervised baseline. The result is a null. No configuration in the grid of 18 runs credibly beats supervision on tasks the agent has already largely mastered, and once the learning rate leaves a narrow low band the agent gets worse. The null is also robust. It survives adding evaluation and training seeds, adding the schedule and group-size choices the grid omitted, switching from a text observation to a Set-of-Marks screenshot, and scaling the backbone to 8B. Figure~\ref{fig:teaser} previews the study and its outcome.

One reading of a null is that the pipeline is broken or underpowered and would fail to climb anywhere. A positive control rules this out. Using the identical harness, reward, and recipe, we select tasks whose reward is reachable by sampling, where the model already succeeds under temperature more often than under greedy decoding, and retrain. There the agent climbs by up to 22 points, with a paired interval that excludes zero. GRPO improves the agent when the sampled policy can already beat the greedy one, and it stalls when greedy is the better of the two, which is the regime a competent supervised agent occupies. A practitioner can act on this by measuring the gap between sampled and greedy success first and expecting little from reinforcement learning when that gap is not positive. For a competent small web agent, extra skill comes from supervision or distillation, not from more reinforcement learning.

We then open the model to ask why the harmful regimes fail. A low learning rate leaves an update that is real but too small to change behavior; a middle rate degrades the agent, and a high rate collapses it. The two failures are different lesions. The degrade regime destroys the effective rank \citep{roy2007effective} of the late layers while leaving earlier layers intact, whereas the collapse regime holds that rank at or above supervision and instead destroys the readout, an inversion we confirm at nearly two thousand fixed states. To test whether this correlation is causal, we graft weights. Restoring the attention or MLP blocks from initialization repairs the degraded agent, while restoring the embedding, whose drift dominates the raw weight change, does nothing, which points to the compute blocks as the site of the failure and clears the embedding despite its large movement \citep{ilharco2022editing}. Read in the other direction, the same probe shows late-layer rank rising with capability where the agent climbs and falling where it degrades, so at 4B rank tracks skill both ways. At 8B this coupling breaks down, and we say so.

Our choice of a single benchmark is deliberate and follows from the interpretability goal. A mechanistic account requires holding everything except the intervention fixed, so that a change in effective rank or success rate can be attributed to the learning rate and not to uncontrolled variation in the environment. MiniWoB provides that control. Its tasks are deterministic under a seed, its success check is exact, and the breadth of interface skills it covers, from clicking and typing to selection and navigation, makes it a broad probe of web-agent behavior, not a narrow one. A benchmark such as WebArena \citep{zhou2024webarena} is closer to deployment but introduces run-to-run variation that is hard to hold constant, and that variation would confound the attributions this study rests on. We treat MiniWoB as the controlled setting in which the mechanism can be read cleanly, and we are clear about what that costs in generality.

Our contributions are the following.
\begin{enumerate}
\item A controlled null: across a grid of 18 runs and several robustness checks, GRPO does not credibly add skill to a competent small language and vision-language model web agent, and higher learning rates credibly remove it.
\item A positive control that identifies the cause of the null as a headroom condition, together with a cheap test of sampled against greedy success that can be applied beforehand.
\item A mechanistic account of the failure as a learning-rate-gated double dissociation, causally localized by weight grafting, with the dominant embedding drift shown to be a correlational red herring.
\item Evidence that late-layer effective rank tracks capability in both directions at 4B and dissociates at 8B, reported as a property that depends on scale.
\item A full battery of paired statistics, equivalence tests, and interpretability measurements, released so others can check the null for themselves.
\end{enumerate}

\section{Related Work}

\begin{table*}[!t]
\centering
\small
\setlength{\tabcolsep}{6pt}
\begin{tabular}{@{}c l p{0.40\textwidth} p{0.30\textwidth}@{}}
\toprule
 & Experiment & Principle & Purpose \\
\midrule
A & Controlled null (\S\ref{sec:setup:grpo}, \S\ref{sec:setup:eval}) & An 18-run grid over learning rate, KL weight, seed, initialization, and clip bound on one fixed harness, each arm paired against the supervised baseline (Eq.~\ref{eq:advantage}--\ref{eq:mcnemar}). & Decide whether any GRPO configuration credibly improves the agent's success rate. \\
\addlinespace
B & Positive control (\S\ref{sec:setup:interp}) & The identical harness, reward, and recipe on tasks where sampled success already exceeds greedy (Eq.~\ref{eq:head}). & Separate a genuine null from a broken pipeline and identify the headroom condition. \\
\addlinespace
C & Mechanism (\S\ref{sec:setup:interp}) & Fixed-state rank and output-behavior probes, plus causal weight grafting on the trained checkpoints (Eq.~\ref{eq:erank}--\ref{eq:graft}). & Explain the failure at the level of individual weight groups. \\
\bottomrule
\end{tabular}
\caption{The three experiments of this study, the principle behind each, and what each is for. Section~\ref{sec:setup} specifies the methods and statistics; the results follow in later sections.}
\label{tab:experiments}
\end{table*}

Reinforcement learning with verifiable rewards has driven recent gains in reasoning models, and GRPO \citep{shao2024deepseekmath}, with refinements such as decoupled clipping \citep{yu2026dapo} and less biased advantage estimation \citep{liu2025understanding} on top of the proximal policy optimization objective \citep{schulman2017proximal}, is now a standard recipe. Whether these gains reflect genuinely new capability or a sharpening of behavior the base model can already produce is contested, with evidence that reinforcement learning often fails to extend a model beyond what sampling already reaches \citep{chen2026does}. That question has been argued mostly on reasoning benchmarks through pass@k. We bring it to an interactive agent and ask it as a practitioner would, against the agent's own supervised baseline.

For web agents, both supervised and reinforcement learning pipelines have raised success rates across interactive benchmarks, from the MiniWoB environment \citep{shi2017world,liu2018reinforcement} to the larger WebArena \citep{zhou2024webarena}, and online curriculum methods report further gains \citep{qi2025webrl}. This line of work shows that reinforcement learning can help a web agent. It does not settle whether it helps a small language and vision-language model agent on tasks the agent has already learned to solve, which is the regime a competent deployed system usually occupies and the one we isolate here.

Our mechanistic analysis builds on tools for reading a model's internal state. Effective rank \citep{roy2007effective} measures how many directions a representation actually uses, and editing in weight space shows that behavior can be moved by recombining parameter groups \citep{ilharco2022editing}. We turn these into a causal test rather than a correlational one: we restore a single component group from its initialization, keep the rest, and re-measure success, which separates the parameters that carry a failure from those that merely move the most.

Two gaps motivate this paper. The sharpening question has not been settled for an agent by a controlled comparison against its own supervised start with a matched positive control, and where a reinforcement learning null is reported it is seldom explained at the level of the weights. We address both. We establish a controlled null for a small language and vision-language model web agent, use a positive control to show the null is a property of the task rather than the training pipeline, and give a causal account of the failure at the level of individual weight groups. To our knowledge this is the first study to pair a controlled reinforcement learning null for a web agent with a mechanism for why it occurs.

\section{Experimental Setup}
\label{sec:setup}

Our aim is to decide whether GRPO adds skill to a competent small language and vision-language web agent and, when it does not, to explain why. The design follows directly. We fix a single agent, environment, reward, and task set, so that any change in success rate traces to the reinforcement-learning choices and not to the setup, and on that fixed harness we run the three experiments summarized in Table~\ref{tab:experiments}. The rest of this section specifies the agent and reward (\S\ref{sec:setup:agent}), the GRPO recipe and grid (\S\ref{sec:setup:grpo}), the evaluation and statistics that make the null credible (\S\ref{sec:setup:eval}), and the interpretability probes (\S\ref{sec:setup:interp}).

\subsection{Agent, Environment, and Reward}
\label{sec:setup:agent}

The agent is Qwen3-VL at 4B and 8B \citep{bai2025qwen3}, run on MiniWoB \citep{shi2017world,liu2018reinforcement}. At each step the policy $\pi_\theta$ reads an observation and emits one action from a compact vocabulary of clicks, typing, selection, navigation, and an explicit \texttt{finish}, up to a fixed step budget. Two observation tracks share all other logic: a \emph{text} track that serializes the page, and a \emph{Set-of-Marks} track \citep{yang2023set} that presents the screenshot with numbered marks. The control grid and the mechanism use the text track; the Set-of-Marks track is a modality check. The reward is a sparse binary terminal signal, broadcast to every step of the episode,
\begin{equation}
r(\tau)=\mathbf{1}\!\left[\operatorname{success}(\tau)\right]\in\{0,1\}.
\end{equation}

\subsection{GRPO Recipe and Control Grid}
\label{sec:setup:grpo}

For each training prompt the policy samples a group of $G$ rollouts. Following \citet{liu2025understanding}, the advantage is the group mean-centered reward, with no division by the group standard deviation,
\begin{equation}
\label{eq:advantage}
A_i = r_i-\frac{1}{G}\sum_{j=1}^{G}r_j.
\end{equation}
The per-token importance ratio between the current policy and the policy that produced the rollout is
\begin{equation}
\label{eq:ratio}
\rho_c(\theta)=\frac{\pi_\theta(y_c\mid x,y_{<c})}{\pi_{\theta_{\mathrm{old}}}(y_c\mid x,y_{<c})},
\end{equation}
and the clipped surrogate minimized over a segment of $C$ completion tokens is
\begin{equation}
\label{eq:grpo}
\mathcal{L}_{\mathrm{GRPO}}=-\frac{1}{C}\sum_{c=1}^{C}\min\!\Big(\rho_c A,\ \operatorname{clip}(\rho_c,1-\epsilon_{\mathrm{lo}},1+\epsilon_{\mathrm{hi}})\,A\Big),
\end{equation}
with an asymmetric clip-higher bound \citep{yu2026dapo}, $\epsilon_{\mathrm{lo}}=0.20<\epsilon_{\mathrm{hi}}=0.28$. An optional KL anchor adds $\beta\widehat{D}_{\mathrm{KL}}$ to the loss, using the non-negative $k_3$ estimator against the frozen initialization $\pi_{\mathrm{ref}}$ with $u_c=\log(\pi_{\mathrm{ref}}/\pi_\theta)$,
\begin{equation}
\label{eq:kl}
\widehat{D}_{\mathrm{KL}}=\frac{1}{C}\sum_{c}\big(e^{u_c}-u_c-1\big).
\end{equation}
The learning rate follows a linear warmup over a fraction $f$ of the $R$ rounds, then a cosine decay,
\begin{equation}
\label{eq:lr}
\eta_t=\eta_0\cdot\begin{cases}(t+1)/w,& t<w,\\[3pt]\tfrac12\big(1+\cos\pi\tfrac{t-w}{R-w}\big),& t\ge w,\end{cases}\qquad w=\lceil fR\rceil.
\end{equation}
The controlled null (experiment~A) is a grid of 18 runs varying learning rate, KL weight $\beta$, seed, initialization, and the clip bound, with a recipe-ablation battery that adds the warmup and cosine schedule and the group size $G\in\{8,16,32\}$.

\subsection{Evaluation Protocol and Statistics}
\label{sec:setup:eval}

Every run is scored with clean greedy decoding on the set $\mathcal{E}$ of 11 tasks at 5 seeds, giving 55 matched episodes. The success rate over $\mathcal{E}$ is
\begin{equation}
\label{eq:sr}
\mathrm{SR}=\frac{1}{|\mathcal{E}|}\sum_{e\in\mathcal{E}}\mathbf{1}[\operatorname{success}(e)],
\end{equation}
reported at level $z$ with the Wilson score interval \citep{wilson1927probable}
\begin{equation}
\label{eq:wilson}
\frac{\widehat p+\tfrac{z^2}{2n}\pm z\sqrt{\tfrac{\widehat p(1-\widehat p)}{n}+\tfrac{z^2}{4n^2}}}{1+z^2/n}.
\end{equation}
Because a MiniWoB reward is deterministic given the task seed, each $(\text{task},\text{seed})$ pair is matched across arms and every arm is compared to the supervised baseline with a paired test. On the discordant pairs $(b,c)$ the exact two-sided McNemar $p$-value is \citep{mcnemar1947note}
\begin{equation}
\label{eq:mcnemar}
p=2\!\!\sum_{i=0}^{\min(b,c)}\!\binom{b+c}{i}2^{-(b+c)},
\end{equation}
and a task-clustered bootstrap resamples the 11 task clusters to give a $95\%$ interval on the success-rate difference $\Delta=\mathrm{SR}_{\mathrm{arm}}-\mathrm{SR}_{\mathrm{SFT}}$. We call an arm \emph{credibly better} or \emph{credibly worse} only when this interval excludes zero; a positive point estimate whose interval touches zero is no credible difference. Equivalence within a margin $\delta$ is a two one-sided test \citep{schuirmann1987comparison}: the $90\%$ interval of $\Delta$ lies inside $[-\delta,+\delta]$.

\subsection{Interpretability Measurements}
\label{sec:setup:interp}

On a fixed cache of $N$ hidden states we read a small set of quantities. The effective rank of a layer is the exponential of the entropy of its normalized covariance eigenvalues $\{\lambda_k\}$ \citep{roy2007effective}, and we also report the participation ratio and the stable rank of the same state matrix $X$,
\begin{equation}
\label{eq:erank}
\operatorname{erank}=\exp\!\Big(\!-\!\sum_k p_k\log p_k\Big),\quad p_k=\frac{\lambda_k}{\sum_j\lambda_j},
\end{equation}
\begin{equation}
\label{eq:pr}
\mathrm{PR}=\frac{\big(\sum_k\lambda_k\big)^2}{\sum_k\lambda_k^2},\qquad \mathrm{sr}=\frac{\lVert X\rVert_F^2}{\lVert X\rVert_2^2}.
\end{equation}
Three probes read the output distribution $p_n=\operatorname{softmax}(\ell_n)$. Argmax agreement with the initialization counts how often the top next-token prediction is unchanged,
\begin{equation}
\label{eq:agr}
\operatorname{agr}=\frac{1}{N}\sum_{n}\mathbf{1}\!\left[\arg\max\ell^{\theta}_n=\arg\max\ell^{\mathrm{ref}}_n\right],
\end{equation}
while the mean entropy and the degeneracy, the fraction of states whose mass collapses onto one token above a threshold $\tau$, are
\begin{equation}
\label{eq:beh}
\begin{aligned}
H&=\frac{1}{N}\sum_{n}\Big(\!-\!\sum_v p_{n,v}\log p_{n,v}\Big),\\
\operatorname{deg}&=\frac{1}{N}\sum_{n}\mathbf{1}\!\big[\textstyle\max_v p_{n,v}>\tau\big].
\end{aligned}
\end{equation}
A causal graft restores one component group $\mathcal{G}$, attention, MLP, or embedding, from initialization and re-evaluates,
\begin{equation}
\label{eq:graft}
\theta'=\theta,\quad \theta'_{\mathcal{G}}\leftarrow\theta^{\mathrm{init}}_{\mathcal{G}},\qquad \text{recovery}=\mathrm{SR}(\theta').
\end{equation}
Finally, the headroom of a task is the gap between the sampled and greedy success of the supervised policy,
\begin{equation}
\label{eq:head}
\Delta_{\mathrm{head}}=\mathrm{SR}_{\mathrm{sample}}-\mathrm{SR}_{\mathrm{greedy}}.
\end{equation}
GRPO can climb only when $\Delta_{\mathrm{head}}>0$; the positive control (experiment~B) selects tasks on this criterion, and the grafts drive the mechanism analysis (experiment~C).

\section{When GRPO Fails}
\label{sec:null}

We begin with the regime a deployed agent occupies, the tasks it has already learned to solve. The supervised policy clears the 11-task grid at a success rate of 49.1\% (27 of 55 matched episodes), an aggregate that mixes tasks it solves almost every time with a harder frontier. The question is whether any reinforcement learning configuration moves that number upward, and none does. The two strongest arms are nominally above supervision by 3.6 points, yet their paired intervals stretch from zero to about ten points and the exact McNemar test returns $p$ between 0.50 and 0.69, so the nudge is indistinguishable from noise (Table~\ref{tab:regimes}). Nor does the result hinge on the learning rate alone. Varying the KL weight, the initialization, and the clip bound each moves the point estimate by at most two points and none of them credibly, so no axis of the 18-run grid turns into a gain. Even taking the best arm as the headline, an upward-biased choice, leaves an estimate that only matches supervision plus a statistically empty 3.6 points.

A flat result invites the reading that the tasks leave no room to improve, but they do. Restricting the score to the frontier subset, defined as the tasks whose supervised success sits between 0.2 and 0.8, the baseline is 37.1\% and the best configuration reaches 42.9\%, a gain whose paired interval still includes zero, so the headroom that exists is not credibly converted into gain. Stated the other way, two one-sided tests place four of the five strongest configurations within five points of supervision, one of them episode for episode identical, which reads the outcome as an equivalence: on the mastered tasks GRPO matches the supervised policy but does not improve on it.

\begin{figure*}[t]
\centering
\includegraphics[width=0.32\textwidth]{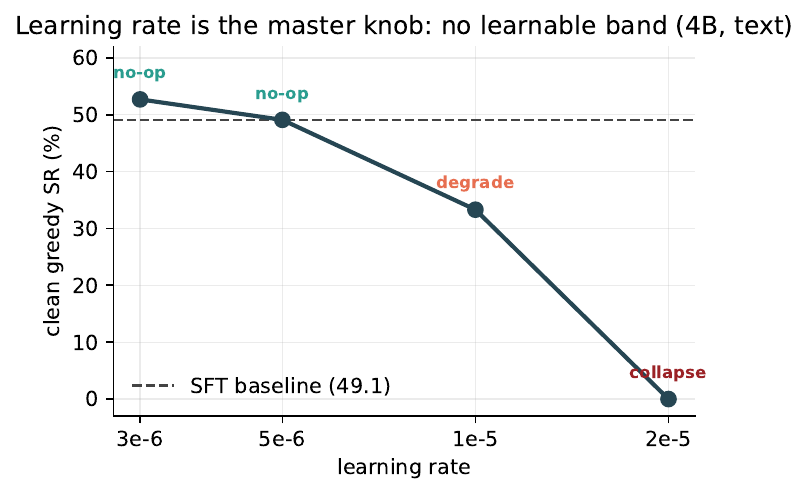}\hfill
\includegraphics[width=0.32\textwidth]{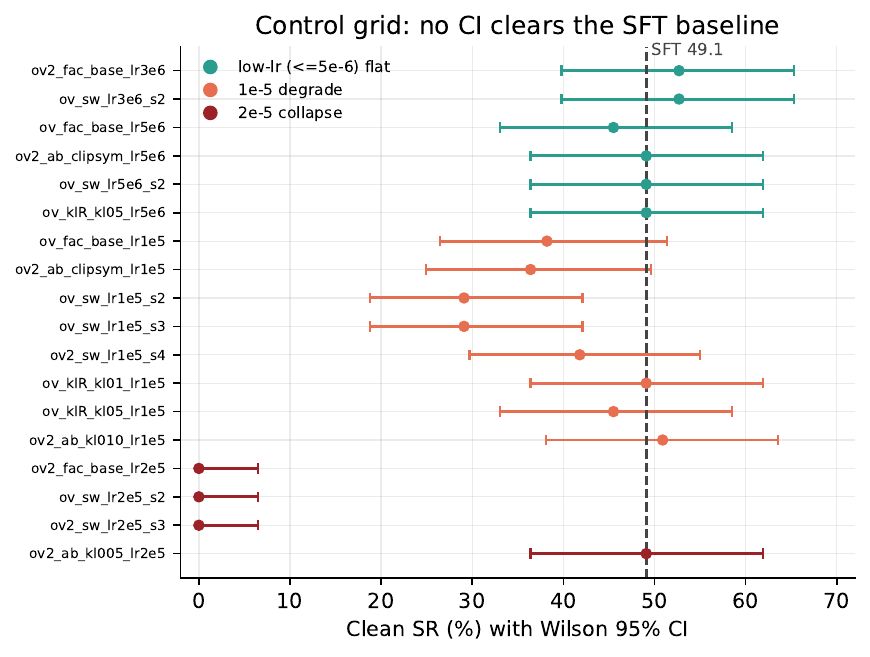}\hfill
\includegraphics[width=0.32\textwidth]{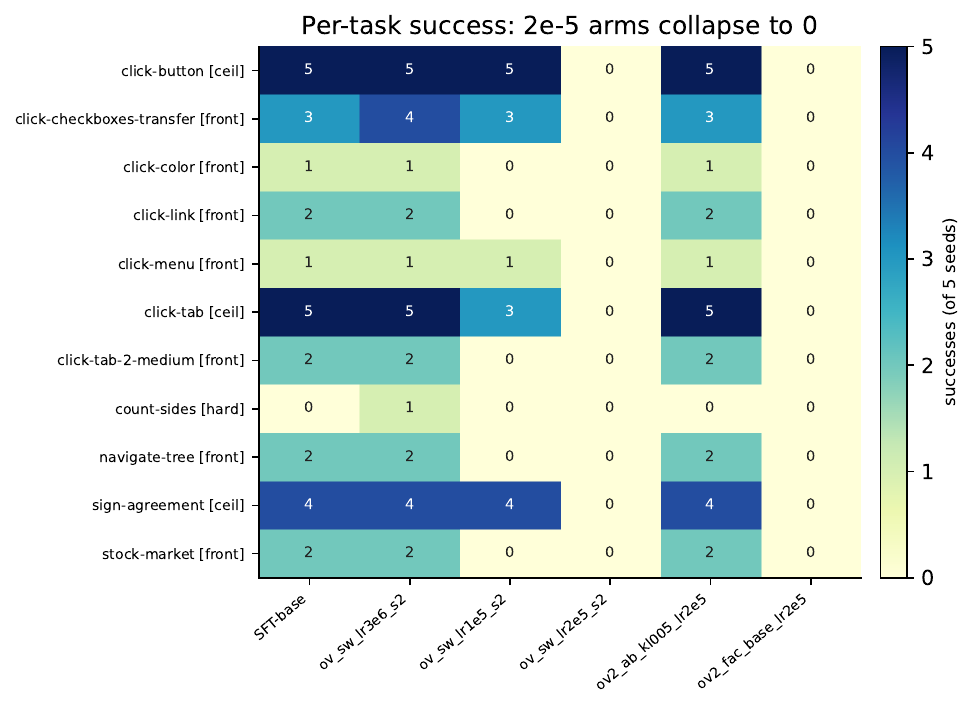}
\caption{The null across the grid (4B, text). Left: clean greedy success against learning rate, flat at the low rates where the update is a functional no-op, then degrading and collapsing as the step grows, with no setting clearing the supervised baseline (dashed). Middle: a forest plot of all 18 arms with Wilson intervals, none of which clears the baseline. Right: the per-task success matrix, where the moderate and high rates drive whole tasks to zero while the low-rate arm tracks supervision.}
\label{fig:lr}
\end{figure*}

The one variable that moves success is the learning rate, and it moves it only downward. Holding initialization and the KL weight fixed on the text track, clean success falls as the step size grows, from 52.7\% at $3\times10^{-6}$, through the supervised baseline near $5\times10^{-6}$, down to 33.3\% at $1\times10^{-5}$ and zero at $2\times10^{-5}$ (Figure~\ref{fig:lr}). Three regimes fall out. At the low end the update is a functional no-op that leaves behavior at supervision. A middle rate degrades the agent by a credible 15 points, and the highest rate collapses it to zero, both with intervals that exclude the baseline and McNemar $p$ below 0.04 and 0.001 (Table~\ref{tab:regimes}). The only rates that leave the agent unharmed are the ones too small to change behavior, and no rate above that band clears the baseline. This monotone shape is specific to the 4B text track; on the Set-of-Marks track and at 8B the nominal peak sits elsewhere, so we keep the shape scoped to 4B text.

The null does not rest on any single choice in the grid, and the full battery is collected in Appendices~\ref{app:robust} and~\ref{app:track}. Adding 25 evaluation seeds lowers the baseline to 44.7\% and still surfaces no credible winner; the one arm that reaches $p=0.001$ under an unclustered test loses that status once episodes are clustered by task, and a replication over 6 training seeds averages 49.7\%, back on the baseline, so its apparent edge was a lucky training seed. The warmup and cosine schedule the grid omitted remove the collapse pathology, since moderate rates now dip and recover instead of dying, but a recovery is not a gain and the schedule still produces no credible improvement. A larger sampling group of 16 or 32 does not help, and at a high rate it breaks the agent sooner, since a larger group is a larger effective step. The Set-of-Marks screenshot raises the baseline to 63.6\% and again yields no credible gain, and the 8B backbone repeats the pattern, a nominal low-rate nudge that never clears the interval and credibly worse success at the higher rates. The null holds across every axis a practitioner would turn.

A null could also be an artifact of reward-hacking or of scoring a checkpoint after it broke, and neither is at work here. If the agent were gaming the sparse reward by finishing early, the flat runs would fill with premature terminations, but the per-episode taxonomy puts reward-hacking near zero in every regime, and the degrade and collapse runs fail because they emit invalid output, not because they stop short. If the null came from scoring the final checkpoint after it had already broken, then selecting the best checkpoint by held-out success would surface a hidden win. It does not. Scoring every saved round and running a paired test on the best one, the strongest flat run still gains only 3.6 points with an interval that touches zero, and the brief early rise of the degrade run falls apart before it is credible. The null comes from the setting itself: it is not an artifact of when we stopped, and it is not the agent gaming the reward. That still leaves the worry that the pipeline cannot climb anywhere, which the next section settles.

\begin{table}[t]
\centering
\small
\setlength{\tabcolsep}{4.5pt}
\fit{\begin{tabular}{@{}l c c c c@{}}
\toprule
Regime & LR & SR & $\Delta$SR & McNemar $p$ \\
\midrule
no-op & $3\times10^{-6}$ & 52.7 & $+3.6\ [+0.0,+10.9]$ & 0.50 \\
baseline & -- & 49.1 & -- & -- \\
degrade & $1\times10^{-5}$ & 33.3 & $-15.0\ [-20.9,-9.5]$ & ${\le}0.04$ \\
collapse & $2\times10^{-5}$ & 0.0 & $-49.1\ [-62.7,-36.4]$ & ${<}0.001$ \\
\bottomrule
\end{tabular}}
\caption{Learning rate is the only dial that moves success, and only downward (4B, text track, supervised initialization, no KL). The degrade and collapse rows pool the three and two seeds run at those rates, and $\Delta$SR is the task-clustered paired difference against the supervised baseline, so for a pooled cell it need not equal the difference of the marginal rates. The low-rate no-op is indistinguishable from supervision; the moderate and high rates are credibly worse.}
\label{tab:regimes}
\end{table}

\section{When GRPO Succeeds}
\label{sec:pc}

The previous section ended on the obvious worry that the pipeline simply cannot climb anywhere. We test it by keeping the harness, reward, and recipe exactly as before, still on the 4B text track, and changing only the tasks. We profiled the supervised policy on all 119 MiniWoB tasks, recording both its greedy success and the success it reaches under temperature sampling, and selected ten tasks with reachable headroom, where sampling already succeeds more often than greedy and there is therefore a policy for GRPO to move toward that greedy does not already reach. Two examples fix the idea: drawing a line, where greedy never succeeds but sampling does more than half the time, and filling an autocomplete field, where sampling roughly doubles the greedy rate.

\begin{figure}[t]
\centering
\includegraphics[width=\columnwidth]{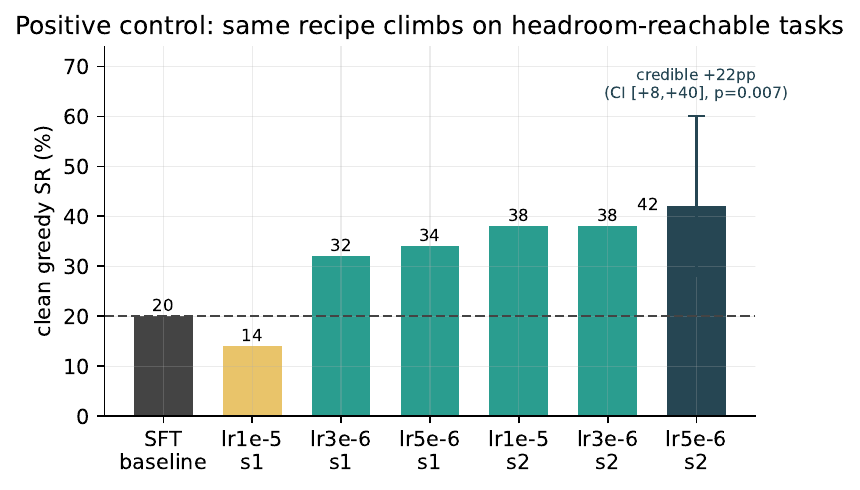}
\caption{Positive control on the headroom set. The supervised baseline (dashed) solves the ten tasks at 20\%; five of six configurations climb, and the best reaches 42\% with a paired interval that excludes zero. The low rate that does nothing on the mastered grid climbs here.}
\label{fig:pc}
\end{figure}

On this set the same recipe climbs. Scored on the ten tasks at 5 seeds, 50 matched episodes, the supervised baseline solves 20.0\%, and the strongest configuration reaches 42.0\%, a gain of 22 points whose task-clustered interval runs from 8 to 40 and excludes zero, with McNemar $p=0.007$ (Figure~\ref{fig:pc}). The climb is not one lucky run. Five of the six configurations rise between 12 and 22 points, four of them credibly, and only a single run at a middle rate degrades. The low learning rate that was a pure no-op on the mastered grid gains 12 to 18 points here, so the variable that decides the outcome is the task and not the step size.

Table~\ref{tab:contrast} puts the two settings side by side. The recipe that cannot separate from supervision on the mastered grid produces a credible 22-point gain on the headroom set, so the null is not a broken or underpowered pipeline; it is GRPO failing where a working GRPO would also fail, on tasks whose reward the greedy policy already reaches. This explains the null in terms of the objective itself. GRPO optimizes sampled rollouts, so it can only climb toward a policy that sampling reaches. On the mastered grid the gap runs the wrong way: greedy is at 0.49 against a sampled 0.32 to 0.38, so there is nothing above greedy to move toward. On the headroom set the gap is positive by construction, and the agent climbs. The gap of Eq.~\ref{eq:head} is therefore a cheap screen a practitioner can run before committing compute, since it predicts both the null and the climb from the supervised policy alone.

\begin{table}[t]
\centering
\small
\setlength{\tabcolsep}{5pt}
\fit{\begin{tabular}{@{}l c c c c@{}}
\toprule
Task set & SFT & Best & $\Delta$SR & McNemar $p$ \\
\midrule
Mastered grid & 49.1 & 52.7 & $+3.6$ & 0.50 \\
Headroom set & 20.0 & 42.0 & $+22.0$ & 0.007 \\
\bottomrule
\end{tabular}}
\caption{The same harness, reward, and recipe on two task sets, the mastered 11-task grid and the 10-task headroom set, each at 5 seeds. On the mastered grid the best arm gains only $+3.6$ points, with a task-clustered $95\%$ interval $[+0.0,+10.9]$ that includes zero. On the headroom set, where sampling already beats greedy, it gains a credible $+22.0$ points, interval $[+8.0,+40.0]$. The null is a property of the task, not of the pipeline.}
\label{tab:contrast}
\end{table}

\section{Why GRPO Fails}
\label{sec:mech}

\begin{figure*}[t]
\centering
\includegraphics[width=0.32\textwidth]{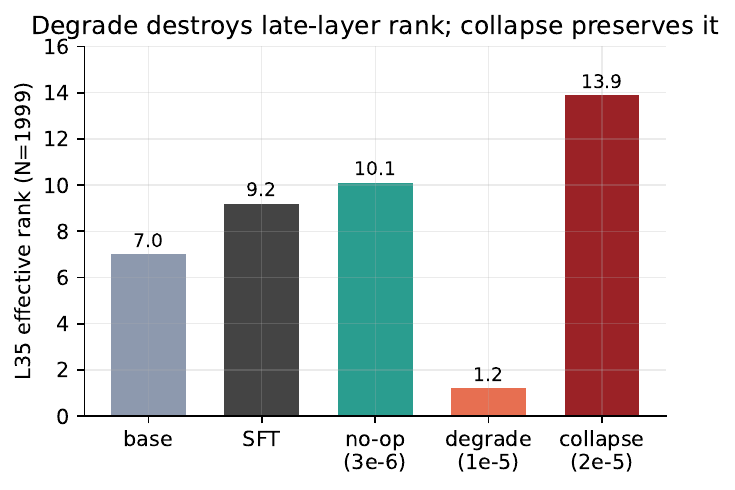}\hfill
\includegraphics[width=0.32\textwidth]{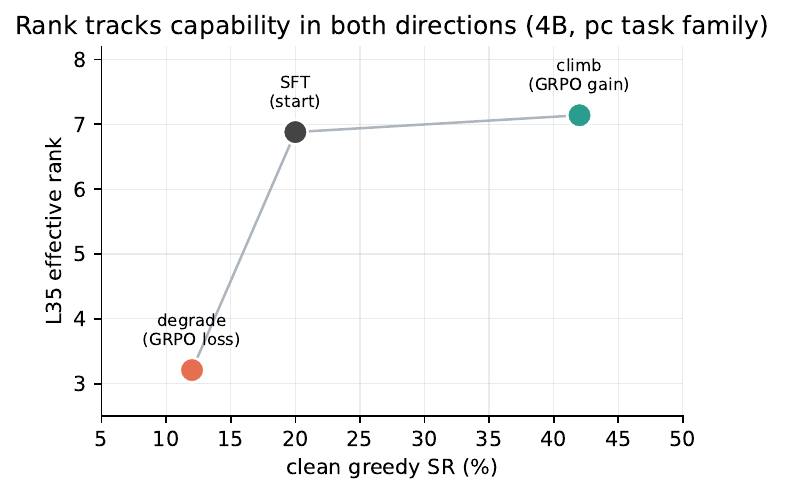}\hfill
\includegraphics[width=0.32\textwidth]{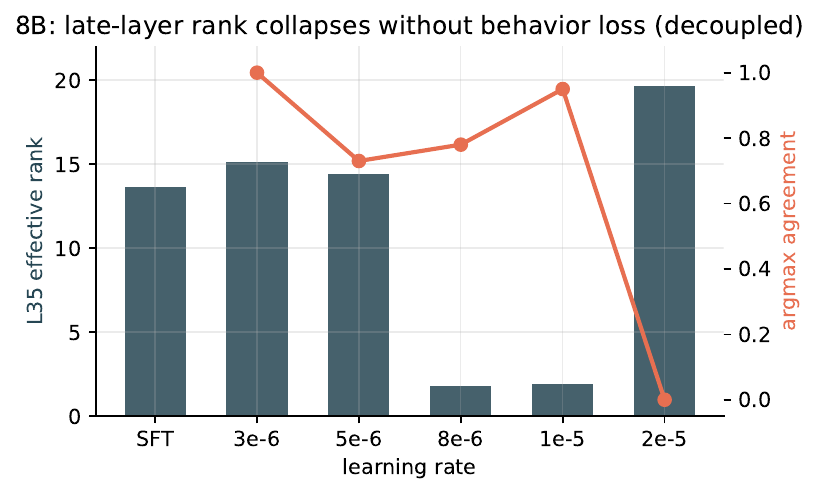}
\caption{The rank account (layer 35). Left: effective rank by regime at 4B, where degrade collapses the rank and collapse preserves or raises it. Middle: on the positive-control family, rank stays healthy where the agent climbs and collapses where it degrades. Right: at 8B, rank and argmax agreement never fall together across the learning-rate sweep, so the coupling does not transfer. Full probes, including argmax agreement and degeneracy by regime, are in Appendix~\ref{app:mech}.}
\label{fig:rankstory}
\end{figure*}

The headroom condition says when GRPO fails to help, but not how the harmful learning rates damage the model, so we open it up. Reading a small set of quantities on a fixed cache of hidden states shows that the two harmful regimes damage the model in different ways. The middle rate that degrades the agent destroys the effective rank of the late layers, read at layer 35, which falls from about nine at supervision to near one while the earlier layers keep their rank. The high rate that collapses the agent leaves that late-layer rank at or above supervision yet drives the argmax agreement with initialization to zero (Figure~\ref{fig:rankstory}, left). Both regimes push the readout down, collapse completely and degrade in part, but only degrade also destroys the representation. This rank inversion is stable across the full cache of 1999 states, so it is not an artifact of thin sampling.

Rank and agreement are correlations, and to make them causal we graft. Restoring one component group from initialization into the trained model, keeping the rest, and re-scoring isolates the weights that carry the failure. For the degraded agent, restoring the attention or the MLP alone lifts frontier success from 11\% to 37\% and 40\%, at or just above the supervised level, while restoring the embedding leaves it near 14\% (Figure~\ref{fig:graft}; the full recovery table for both regimes is in Appendix~\ref{app:mech}). Either compute pathway is sufficient, so the failure is localizable, and the direction holds across three degrade checkpoints. This is where a correlational reading would have gone wrong. The embedding drifts more than any other group in the raw weight difference, so an argument from movement alone would blame it, yet the graft shows the embedding is causally inert while the attention and MLP carry the damage. To rule out a generic undoing of drift, we restore a random group of matched parameter count as a null; across two dozen draws it recovers to a mean of 19\% and a 95th percentile of 23\%, and the real attention and MLP grafts sit well above that band, so the localization is specific and not a side effect of moving weights back toward their start.

\begin{figure}[t]
\centering
\includegraphics[width=\columnwidth]{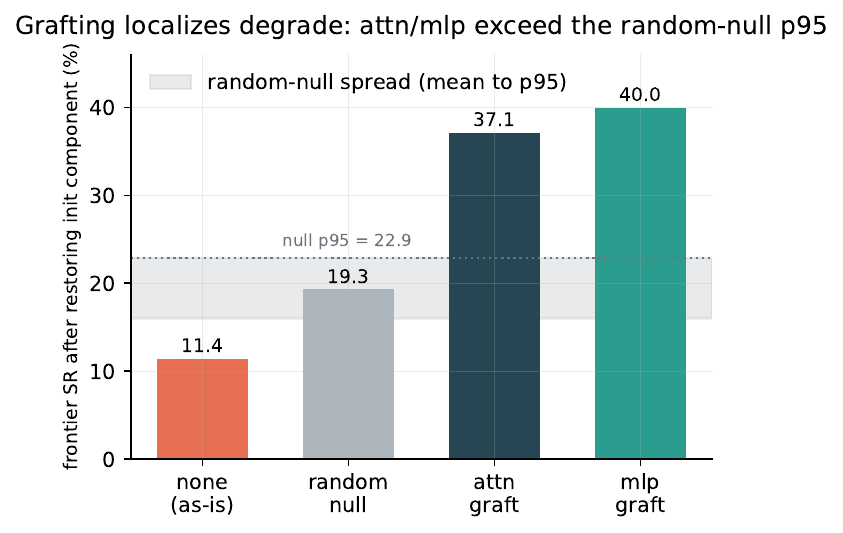}
\caption{Causal grafting on the degraded agent. Restoring the attention or MLP from initialization recovers frontier success past the random-null band (its mean to 95th percentile), while restoring the embedding does not, so the damage is localized to the compute pathways and not to the group that moved most.}
\label{fig:graft}
\end{figure}

The collapsed agent behaves differently under the same test. No single group restores it, attention and MLP alone recover almost nothing, and only restoring every group together returns it to the supervised level, so collapse is a distributed corruption of the readout rather than a lesion in one place. This completes a double dissociation: degrade destroys the late-layer rank and is localizable to a single compute pathway, while collapse spares the rank and cannot be localized at all.

For rank to be a meaningful measure, it should rise when the agent improves, not only fall when it breaks. On the family used for the positive control it does. Reading the same late-layer rank on three checkpoints from that family, the supervised start sits at rank 6.9 with 20\% success, the configuration that climbs to 42\% keeps its rank healthy at 7.1, and the configuration that degrades to 12\% has its rank cut to 3.2 (Figure~\ref{fig:rankstory}, middle). The loss shows up as a sharp rank collapse and the gain as a rank that stays healthy rather than falling, so within this family rank health and capability move together, with the caveat that these are single checkpoints and the upward move is small.

This coupling is a property of the 4B model and does not survive a change of scale. Reading the same probe across the learning-rate sweep on the 8B backbone, on the text track, we find no setting where rank and argmax agreement fall together (Figure~\ref{fig:rankstory}, right). At the rates that degrade the 4B model the 8B late-layer rank collapses to about two while the argmax agreement stays between 0.78 and 0.95, so the readout is untouched even though the representation is not, and only at the highest rate does the agreement collapse, with the rank left high. The larger model absorbs a late-layer rank collapse without changing what it predicts. We therefore scope the account: the link between rank and capability holds at 4B, shown in both directions by grafting and by the positive control, and it breaks at 8B. We treat that as a limit of the 4B account.

The mechanism, then, is a pair of learning-rate-gated lesions. A middle rate erases the late-layer rank that the attention and MLP carry, and a high rate erases the readout in a way spread across the whole model. Neither adds anything the supervised agent lacked.

\section{Discussion}
\label{sec:discussion}

The result carries a direct lesson for practice. On a small web agent that already performs its tasks, the next increment of skill does not come from a second stage of reinforcement learning, because the agent's greedy policy already sits above what its own sampling reaches and GRPO can pursue only the sampled policy. The headroom gap of Eq.~\ref{eq:head} is the screen to run first: a value at or below zero on a held-out set is the signal to spend the budget on supervision instead. In a distillation pilot, expert supervision lifted the base model by 30 points on a task set where reinforcement learning on top of any initialization added nothing (Appendix~\ref{app:robust}).

Two boundaries on the claim should be read alongside it. The study is one benchmark and one model family, chosen so the mechanism can be read without the confounds a noisier environment adds, and the robustness battery covers the recipe choices a reviewer would raise (schedule, group size, seeds, observation track, and model scale), none of which changes the verdict. The credible harm from a large step is established on the 4B text track and is only nominal on the Set-of-Marks track, so we state it as a text-track result and not a universal one. The mechanism has its own scope. The link between late-layer rank and capability that grafting and the positive control establish at 4B does not hold at 8B, where a rank collapse and a change in behavior never coincide, so the rank account is a property of the smaller model and not a law across scale.

The finding brings the sharpening debate, so far settled largely on reasoning tasks by pass@k \citep{chen2026does}, to an interactive agent with a controlled answer. Where that debate asks whether reinforcement learning extends a model beyond its base, we ask the same of an agent against its own supervised start and find that it does not, then show with the positive control that the boundary is not the method but the headroom of the task. The contested general claim becomes a specific one we can test: reinforcement learning extends the agent only when its sampled policy already beats its greedy policy.

\section{Conclusion}
\label{sec:conclusion}

We asked whether GRPO adds skill to a small language and vision-language web agent on tasks it has already learned. Across the grid and the robustness checks it does not, and past a narrow band of learning rates it subtracts skill instead. The same pipeline climbs 22 points once the task has headroom, so the null is about the task, not the method. The failure has a clear mechanism: a middle rate erases the late-layer rank that grafting traces to attention and MLP, a high rate erases the readout that no single group repairs, and the dominant embedding drift is inert. For a competent small agent, more skill comes from better supervision, and GRPO earns its compute only where sampling beats greedy.

\section*{Limitations}

This study is deliberately narrow, and several boundaries should be kept in mind. It uses a single benchmark, MiniWoB, and a single model family, Qwen3-VL at 4B and 8B; the null and the mechanism may look different on a more varied environment such as WebArena or on a different backbone. The credible harm from a large learning rate is established on the 4B text track, and on the Set-of-Marks track it is only nominal, so the harm side of the result is text-track evidence and not a universal claim. The mechanism has a scale boundary of its own: the link between late-layer rank and capability holds at 4B but breaks at 8B, so the rank account is specific to the smaller model. The causal grafting and rank-health readings rest on a small number of checkpoints per regime and their point estimates are noisy, so we report them qualitatively. Finally, the null concerns the regime where the agent has already mastered its tasks, and the positive control shows that the same recipe does climb once the task has headroom; we therefore make no claim that reinforcement learning is useless for web agents in general, only that it adds no skill in the competent-agent regime we isolate.

\bibliography{references}

@article{shao2024deepseekmath,
  title={Deepseekmath: Pushing the limits of mathematical reasoning in open language models},
  author={Shao, Zhihong and Wang, Peiyi and Zhu, Qihao and Xu, Runxin and Song, Junxiao and Bi, Xiao and Zhang, Haowei and Zhang, Mingchuan and Li, YK and Wu, Yang and others},
  journal={arXiv preprint arXiv:2402.03300},
  year={2024}
}

@article{yu2026dapo,
  title={Dapo: An open-source llm reinforcement learning system at scale},
  author={Yu, Qiying and Zhang, Zheng and Zhu, Ruofei and Yuan, Yufeng and Zuo, Xiaochen and Yue, Yu and Dai, Weinan and Fan, Tiantian and Liu, Gaohong and Liu, Lingjun and others},
  journal={Advances in Neural Information Processing Systems},
  volume={38},
  pages={113222--113244},
  year={2026}
}

@article{liu2025understanding,
  title={Understanding r1-zero-like training: A critical perspective},
  author={Liu, Zichen and Chen, Changyu and Li, Wenjun and Qi, Penghui and Pang, Tianyu and Du, Chao and Lee, Wee Sun and Lin, Min},
  journal={arXiv preprint arXiv:2503.20783},
  year={2025}
}

@article{schulman2017proximal,
  title={Proximal policy optimization algorithms},
  author={Schulman, John and Wolski, Filip and Dhariwal, Prafulla and Radford, Alec and Klimov, Oleg},
  journal={arXiv preprint arXiv:1707.06347},
  year={2017}
}

@inproceedings{qi2025webrl,
  title={Webrl: Training llm web agents via self-evolving online curriculum reinforcement learning},
  author={Qi, Zehan and Liu, Xiao and Iong, Iat Long and Lai, Hanyu and Sun, Xueqiao and Sun, Jiadai and Yang, Xinyue and Yang, Yu and Yao, Shuntian and Xu, Wei and others},
  booktitle={International Conference on Learning Representations},
  volume={2025},
  pages={79791--79821},
  year={2025}
}

@inproceedings{shi2017world,
  title={World of bits: An open-domain platform for web-based agents},
  author={Shi, Tianlin and Karpathy, Andrej and Fan, Linxi and Hernandez, Jonathan and Liang, Percy},
  booktitle={International Conference on Machine Learning},
  pages={3135--3144},
  year={2017},
  organization={PMLR}
}

@article{liu2018reinforcement,
  title={Reinforcement learning on web interfaces using workflow-guided exploration},
  author={Liu, Evan Zheran and Guu, Kelvin and Pasupat, Panupong and Shi, Tianlin and Liang, Percy},
  journal={arXiv preprint arXiv:1802.08802},
  year={2018}
}

@inproceedings{zhou2024webarena,
  title={Webarena: A realistic web environment for building autonomous agents},
  author={Zhou, Shuyan and Xu, Frank F and Zhu, Hao and Zhou, Xuhui and Lo, Robert and Sridhar, Abishek and Cheng, Xianyi and Ou, Tianyue and Bisk, Yonatan and Fried, Daniel and others},
  booktitle={International Conference on Learning Representations},
  volume={2024},
  pages={15585--15606},
  year={2024}
}

@article{bai2025qwen3,
  title={Qwen3-vl technical report},
  author={Bai, Shuai and Cai, Yuxuan and Chen, Ruizhe and Chen, Keqin and Chen, Xionghui and Cheng, Zesen and Deng, Lianghao and Ding, Wei and Gao, Chang and Ge, Chunjiang and others},
  journal={arXiv preprint arXiv:2511.21631},
  year={2025}
}

@inproceedings{roy2007effective,
  title={The effective rank: A measure of effective dimensionality},
  author={Roy, Olivier and Vetterli, Martin},
  booktitle={2007 15th European signal processing conference},
  pages={606--610},
  year={2007},
  organization={IEEE}
}

@article{ilharco2022editing,
  title={Editing models with task arithmetic},
  author={Ilharco, Gabriel and Ribeiro, Marco Tulio and Wortsman, Mitchell and Gururangan, Suchin and Schmidt, Ludwig and Hajishirzi, Hannaneh and Farhadi, Ali},
  journal={arXiv preprint arXiv:2212.04089},
  year={2022}
}

@article{chen2026does,
  title={Does reinforcement learning really incentivize reasoning capacity in llms beyond the base model?},
  author={Chen, Zhiqi and Lu, Rui and Zhao, Andrew and Wang, Zhaokai and Yue, Yang and Song, Shiji and Huang, Gao},
  journal={Advances in Neural Information Processing Systems},
  volume={38},
  pages={57654--57689},
  year={2026}
}

@article{yang2023set,
  title={Set-of-mark prompting unleashes extraordinary visual grounding in gpt-4v},
  author={Yang, Jianwei and Zhang, Hao and Li, Feng and Zou, Xueyan and Li, Chunyuan and Gao, Jianfeng},
  journal={arXiv preprint arXiv:2310.11441},
  year={2023}
}

@article{mcnemar1947note,
  title={Note on the sampling error of the difference between correlated proportions or percentages},
  author={McNemar, Quinn},
  journal={Psychometrika},
  volume={12},
  number={2},
  pages={153--157},
  year={1947},
  publisher={Springer-Verlag}
}

@article{schuirmann1987comparison,
  title={A comparison of the two one-sided tests procedure and the power approach for assessing the equivalence of average bioavailability},
  author={Schuirmann, Donald J},
  journal={Journal of pharmacokinetics and biopharmaceutics},
  volume={15},
  number={6},
  pages={657--680},
  year={1987},
  publisher={Springer}
}

@article{wilson1927probable,
  title={Probable inference, the law of succession, and statistical inference},
  author={Wilson, Edwin B},
  journal={Journal of the American Statistical Association},
  volume={22},
  number={158},
  pages={209--212},
  year={1927},
  publisher={Taylor \& Francis}
}

The appendices below collect the full result tables, a gallery of the supporting figures, and the algorithms referenced in the main text. All numbers are transcribed from the same frozen result tables used in the body.

\appendix

\section{The Control Grid in Full}
\label{app:grid}

The controlled null is an 18-run grid on the 4B text track that varies the learning rate, the KL weight, the seed, the initialization (supervised or base), and the clip bound. Every arm is scored by clean greedy decoding on the 11 tasks at 5 seeds, giving 55 matched episodes, and compared to the supervised baseline with the paired test of the main text. Table~\ref{tab:grid} lists the representative arms. Classified by their training dynamics the 18 runs split into 9 functional no-ops, 6 degrade runs, and 3 collapse runs, and no arm's paired interval clears the baseline. Reporting the best of eighteen is itself an upward-biased statistic, and even that best matches supervision plus a statistically empty $3.6$ points. The ``no headroom'' reading is refuted on the frontier subset, the tasks whose supervised success lies in $[0.2,0.8]$: the supervised frontier baseline is $37.1\%$ $[23.2,53.7]$ (13 of 35) and the best configuration reaches only $42.9\%$ $[28.0,59.1]$ (15 of 35), whose interval lower bound of $28.0$ sits below the baseline point estimate.

\begin{table*}[t]
\centering
\small
\setlength{\tabcolsep}{6pt}
\begin{tabular}{@{}l l c c c c c l@{}}
\toprule
Arm & Init & LR & KL $\beta$ & Clean SR [95\%] & $\Delta$SR [95\% CI] & McNemar $p$ & Verdict \\
\midrule
Supervised baseline & SFT & --- & 0 & 49.1 [36.4, 61.9] & --- & --- & baseline \\
Base initialization & base & --- & 0 & 50.9 [38.1, 63.6] & $+1.8$ & --- & no-diff \\
Best, SFT init & SFT & $3\times10^{-6}$ & 0 & 52.7 [39.8, 65.3] & $+3.6\ [+0.0,+10.9]$ & 0.50 & no-diff \\
Best, base init & base & $3\times10^{-6}$ & 0 & 52.7 [39.8, 65.3] & $+3.6\ [+0.0,+9.1]$ & 0.69 & no-diff \\
Low rate & SFT & $5\times10^{-6}$ & 0 & 49.1 [36.4, 61.9] & $+0.0$ & --- & no-diff \\
Middle rate, pooled & SFT & $1\times10^{-5}$ & 0 & 33.3 [26.6, 40.8] & $-15.0\ [-20.9,-9.5]$ & ${\le}0.04$ & worse \\
High rate, pooled & SFT & $2\times10^{-5}$ & 0 & 0.0 [0.0, 3.4] & $-49.1\ [-62.7,-36.4]$ & ${<}0.001$ & worse \\
KL rescue & SFT & $2\times10^{-5}$ & 0.05 & 49.1 [36.4, 61.9] & $+0.0$ & --- & no-diff \\
Best KL & SFT & $1\times10^{-5}$ & 0.10 & 50.9 [38.1, 63.6] & $+1.8$ & --- & no-diff \\
\bottomrule
\end{tabular}
\caption{Representative arms of the 18-run control grid (4B, text). Clean SR is greedy success over 55 matched episodes with a Wilson interval; $\Delta$SR is the task-clustered paired difference against the supervised baseline. The middle- and high-rate rows pool 3 and 2 seeds. The KL rescue arm is episode-identical to supervision (discordant pairs $b/c=0/0$). No arm is credibly better than supervision.}
\label{tab:grid}
\end{table*}

\section{Paired Statistics, Equivalence, and Checkpoint Selection}
\label{app:stats}

The verdict uses the exact McNemar test on the discordant pairs and a task-clustered bootstrap on the 11 task clusters, both over episode-matched arms. Beyond the failure to reject, a two one-sided test states a positive equivalence: four of the five strongest configurations lie within $\pm 5$ points of supervision (Table~\ref{tab:tost}), one of them episode-identical. Because the reported success is the final checkpoint, we also retrained the three regime exemplars saving every round and paired-tested the single best-by-eval checkpoint of each; none is credibly superior (Table~\ref{tab:bestckpt}), so the null is not an artifact of scoring an already broken checkpoint.

\begin{table}[t]
\centering
\small
\setlength{\tabcolsep}{5pt}
\fit{\begin{tabular}{@{}l c c@{}}
\toprule
Configuration & $90\%$ CI of $\Delta$ & Equiv.\ ($\pm5$pp) \\
\midrule
Base-init flat & $[+0.0,+4.1]$ & yes \\
KL $0.10$ & $[+0.0,+0.0]$ & yes, identical \\
Recipe, $G{=}16$ & $[-2.7,+4.1]$ & yes \\
Base-init sanity & $[-2.3,+0.9]$ & yes \\
lr $3\times10^{-6}$, s2 & $[+0.0,+15.0]$ & only $\pm15$pp \\
\bottomrule
\end{tabular}}
\caption{Two one-sided equivalence at the 25-seed grain. Four of five top configurations are statistically equivalent to supervision within $\pm5$ points. The lone exception is the lucky-seed arm, whose six-seed replication averages $49.7\%$ (Table~\ref{tab:trainseed}).}
\label{tab:tost}
\end{table}

\begin{table}[t]
\centering
\small
\setlength{\tabcolsep}{4pt}
\fit{\begin{tabular}{@{}l c c c@{}}
\toprule
Regime & Best-ckpt SR & $\Delta$SR [95\% CI] & $p$ \\
\midrule
Flat ($3\times10^{-6}$) & 52.7 & $+3.6\ [+0.0,+9.1]$ & 0.500 \\
Degrade ($1\times10^{-5}$) & ${\sim}51$ (r1) & $+1.8\ [-9.1,+12.7]$ & 1.000 \\
Collapse ($2\times10^{-5}$) & 49.1 (r0) & $-49.1$ final & ${<}0.001$ \\
\bottomrule
\end{tabular}}
\caption{Best-by-eval checkpoint selection. Scoring every saved round and paired-testing the maximum-SR checkpoint of each regime surfaces nominal nudges (flat drifts to $52.7\%$, degrade peaks near $51\%$ before breaking) but none survives the paired test.}
\label{tab:bestckpt}
\end{table}

\section{Robustness Battery}
\label{app:robust}

The null is unchanged by every recipe and evaluation choice a reviewer would raise. Table~\ref{tab:robust} collects the axes. Widening evaluation to 25 seeds lowers the baseline to $44.7\%$ and leaves no credible winner. The one borderline arm (McNemar $p=0.001$ under an unclustered test) loses credibility once episodes are clustered by task, and a six-seed replication of it averages $49.7\%$ (Table~\ref{tab:trainseed}), so the edge was a lucky training seed. Adding the warmup and cosine schedule removes the collapse pathology without producing a gain, and a larger group breaks the agent sooner at high rates (Table~\ref{tab:recipe}). A per-episode taxonomy shows the failures are output degeneration, not reward-hacking: premature finishing is near zero in every regime (Table~\ref{tab:taxonomy}). On a separate distillation pilot (5 tasks, 6 seeds, $n=30$), expert supervision lifts the base model from $50\%$ to $80\%$ while GRPO on top of any initialization adds nothing, which is the evidence that the constructive lever is supervision rather than reinforcement learning.

\begin{table}[t]
\centering
\small
\setlength{\tabcolsep}{4pt}
\fit{\begin{tabular}{@{}l c c l@{}}
\toprule
Robustness axis & Base & Best $\Delta$ & Verdict \\
\midrule
Grid, 11 tasks $\times$ 5 seeds & 49.1 & $+3.6$ & no credible gain \\
Frontier subset & 37.1 & $+5.7$ & no credible gain \\
25 evaluation seeds & 44.7 & $+5.1$ & no credible gain \\
6 training seeds, mean & 49.1 & $+0.6$ & no credible gain \\
Warmup $+$ cosine schedule & 49.1 & $+3.6$ & no credible gain \\
Group size $G\in\{16,32\}$ & 49.1 & -- & no credible gain \\
Set-of-Marks observation & 63.6 & $+7.3$ & no credible gain \\
8B backbone (Set-of-Marks) & 63.6 & $+9.1$ & no credible gain \\
\bottomrule
\end{tabular}}
\caption{The null survives every robustness axis. Base is supervised success and Best $\Delta$ is the largest nominal point-estimate gain of any arm, in percent, except the training-seed row, which reports the six-seed mean. Every listed gain has a paired interval that includes zero.}
\label{tab:robust}
\end{table}

\begin{table}[t]
\centering
\small
\setlength{\tabcolsep}{4pt}
\fit{\begin{tabular}{@{}l c c c l@{}}
\toprule
Schedule / group & LR & SR & $\Delta$SR & Verdict \\
\midrule
Warmup+cosine, $G{=}8$ & $3\times10^{-6}$ & 52.7 & $+3.6$ & no-diff \\
Warmup+cosine, $G{=}8$ & $8\times10^{-6}$ & --- & recover & no-diff \\
Warmup+cosine, $G{=}8$ & $1.5\times10^{-5}$ & --- & recover & no-diff \\
Warmup+cosine, $G{=}8$ & $2\times10^{-5}$ & 0.0 / 32.7 & worse & worse \\
Larger group, $G{=}16$ & $1.5\times10^{-5}$ & 0.0 & collapse & worse \\
\bottomrule
\end{tabular}}
\caption{Recipe ablation. A standard warmup and cosine schedule turns the constant-rate collapse into a dip-and-recover at moderate rates but yields no credible gain, and a larger sampling group is a larger effective step that breaks high rates faster.}
\label{tab:recipe}
\end{table}

\begin{table}[t]
\centering
\small
\setlength{\tabcolsep}{5pt}
\fit{\begin{tabular}{@{}l c c c@{}}
\toprule
Regime & Reward-hack & Invalid output & Correct/valid \\
\midrule
Flat / KL & ${\sim}0\%$ & low & ${\sim}50\%$ / ${\sim}40\%$ \\
Degrade & ${\sim}0\%$ & $63\%$ & rest \\
Collapse & ${\sim}0\%$ & $98\%$ & rest \\
\bottomrule
\end{tabular}}
\caption{Per-episode failure-mode taxonomy. Reward-hacking (premature finish) is near zero in every regime, so the null is not reward gaming; degrade and collapse fail by emitting invalid output while flat and KL preserve the supervised structure.}
\label{tab:taxonomy}
\end{table}

\begin{table}[t]
\centering
\small
\setlength{\tabcolsep}{5pt}
\fit{\begin{tabular}{@{}c c c c c c c@{}}
\toprule
Seed & 1 & 2 & 3 & 4 & 5 & 6 \\
\midrule
Clean SR & 49.1 & 52.7 & 45.5 & 49.1 & 50.9 & 50.9 \\
\bottomrule
\end{tabular}}
\caption{Six-seed replication of the nominal-best configuration (SFT init, constant lr $3\times10^{-6}$, $G=8$). The mean is $49.7\%$ ($\pm2.5$ sd, $95\%$ CI $[47.7,51.7]$), on the supervised baseline of $49.1\%$, so the $+5.1$-point single-seed result was training-seed noise.}
\label{tab:trainseed}
\end{table}

\section{Observation Track and Model Scale}
\label{app:track}

The no-gain null generalizes across observation track and model scale, while the credible harm from a large rate is established on the 4B text track and is only nominal elsewhere. On the Set-of-Marks track the supervised baseline is higher, at $63.6\%$, and no configuration credibly beats it; the high-rate degradation is directional but not credible (Table~\ref{tab:som}). At 8B on Set-of-Marks the baseline is also $63.6\%$, the low-rate nudge is again nominal, and the higher rates are credibly worse (Table~\ref{tab:som8b}). The 8B mechanism cache (372 states, text) reproduces the late-layer rank inversion: supervised $13.6$, flat $15.1$, degrade $1.9$, collapse $19.6$.

\begin{table}[t]
\centering
\small
\setlength{\tabcolsep}{5pt}
\fit{\begin{tabular}{@{}l c c c@{}}
\toprule
Config (4B, SoM) & SR & $\Delta$SR & McNemar $p$ \\
\midrule
Supervised baseline & 63.6 & --- & --- \\
lr $3\times10^{-6}$ & 67.3 & $+3.6$ & no-diff \\
lr $1\times10^{-5}$ & 70.9 & $+7.3$ & no-diff \\
lr $2\times10^{-5}$ & 52.7 & $-10.9$ & 0.070 \\
\bottomrule
\end{tabular}}
\caption{Set-of-Marks track at 4B. No configuration credibly beats the supervised baseline, and the high-rate degradation is nominal ($p=0.070$, task-clustered interval touches zero), not credible as it is on the text track.}
\label{tab:som}
\end{table}

\begin{table}[t]
\centering
\small
\setlength{\tabcolsep}{5pt}
\fit{\begin{tabular}{@{}l c c c@{}}
\toprule
Config (8B, SoM) & SR & $\Delta$SR & McNemar $p$ \\
\midrule
Supervised baseline & 63.6 & --- & --- \\
lr $3\times10^{-6}$ & 72.7 & $+9.1$ & 0.062 \\
lr $1\times10^{-5}$ & 45.5 & $-18.2$ & 0.002 \\
lr $2\times10^{-5}$ & 38.2 & $-25.5$ & 0.001 \\
\bottomrule
\end{tabular}}
\caption{Set-of-Marks track at 8B. The low-rate nudge is nominal and single-seed ($p=0.062$), the same lucky-seed pattern seen at 4B, and the higher rates are credibly worse.}
\label{tab:som8b}
\end{table}

\section{Mechanism in Detail}
\label{app:mech}

The interpretability probes read a fixed cache of hidden states. Movement magnitude does not predict the failure mode: collapse moves less in $L_2$ than degrade yet is more destructive, and the single largest total movement in the grid is the KL rescue arm, which is behaviorally at the baseline (Table~\ref{tab:wmove}). The rank and output-behavior probes give the double dissociation in full, including next-token entropy (Table~\ref{tab:rankfull}). Causal grafting localizes degrade to the attention or MLP and refutes the embedding, whose drift is largest; a magnitude-matched random-component null over 24 draws recovers only to a mean of $19.3\%$ and a 95th percentile of $22.9\%$, below the real grafts, while for collapse no single group recovers (Table~\ref{tab:graft}). The rank-to-capability coupling that holds at 4B does not transfer to 8B, where rank and argmax agreement never fall together across the sweep (Table~\ref{tab:decouple}). The KL anchor is a mode-preserving stabilizer, not a teacher: every KL arm holds argmax agreement at $1.0$ while next-token entropy inflates to $4.9$--$7.7$, and KL suppresses embedding drift (${\sim}0.006$ with KL against ${\sim}0.011$ without, at $1\times10^{-5}$) rather than freezing the weights.

\begin{table}[t]
\centering
\small
\setlength{\tabcolsep}{4pt}
\fit{\begin{tabular}{@{}l c c c@{}}
\toprule
Regime & mean rel.\ $L_2$ & sparse frac.\ & embed drift \\
\midrule
No-op ($3\times10^{-6}$) & 0.00017 & 0.0023--0.0031 & ${\sim}0.0002$ \\
Degrade ($1\times10^{-5}$) & 0.0010--0.0011 & --- & ${\sim}0.011$ \\
Collapse ($2\times10^{-5}$) & 0.0008--0.0010 & up to 0.068 & 0.0002--0.0010 \\
KL rescue ($2\times10^{-5}$) & 0.00198 & 0.115 & ${\sim}0.006$ \\
\bottomrule
\end{tabular}}
\caption{Weight movement by regime. The outcome is not monotone in movement: collapse moves less than degrade yet destroys more, and the KL rescue arm moves the most of all while staying at the supervised baseline.}
\label{tab:wmove}
\end{table}

\begin{table}[t]
\centering
\small
\setlength{\tabcolsep}{4pt}
\fit{\begin{tabular}{@{}l c c c c@{}}
\toprule
Regime & L35 rank & argmax agr & degeneracy & entropy \\
\midrule
base & 7.0 & --- & 0.00 & 0.002--0.011 \\
SFT & 9.2 & 1.00 & 0.00 & ${\sim}0.01$ \\
no-op & 10.1 & 1.00 & 0.00 & 0.078 \\
degrade & 1.2 & 0.07--0.50 & 0.50--0.93 & 7.2--10.3 \\
collapse & 13.9 & 0.00 & 0.80--1.00 & 0.17--9.04 \\
\bottomrule
\end{tabular}}
\caption{Full fixed-state probes at $N=1999$ (4B, text). Degrade destroys the late-layer rank while collapse preserves or raises it; both drive the readout down, collapse completely and degrade in part. Entropy is not monotone across regimes, so entropy inflation alone is harmless under greedy decoding.}
\label{tab:rankfull}
\end{table}

\begin{table}[t]
\centering
\small
\setlength{\tabcolsep}{7pt}
\fit{\begin{tabular}{@{}l c c@{}}
\toprule
Restored group & degrade & collapse \\
\midrule
none (as trained) & 11.4 & 0.0 \\
random null (mean) & 19.3 & 0.0 \\
random null (95th pct.) & 22.9 & --- \\
embedding & 14.0 & 0.0 \\
attention & 37.1 & 0.0 \\
MLP & 40.0 & 2.9 \\
all groups & 37.1 & 37.1 \\
\bottomrule
\end{tabular}}
\caption{Frontier success (\%) after restoring a component group from initialization. For degrade, attention or MLP alone recovers above the random-null 95th percentile, and the direction holds across three degrade checkpoints (recovery $33$--$111\%$ of the none-to-all gap). For collapse no single group recovers. Values are over 35 episodes and noisy in the point estimate.}
\label{tab:graft}
\end{table}

\begin{table}[t]
\centering
\small
\setlength{\tabcolsep}{7pt}
\fit{\begin{tabular}{@{}l c c@{}}
\toprule
LR (8B, text) & L35 rank & argmax agr \\
\midrule
SFT & 13.63 & --- \\
$3\times10^{-6}$ & 15.14 & 1.00 \\
$5\times10^{-6}$ & 14.39 & 0.73 \\
$8\times10^{-6}$ & 1.80 & 0.78 \\
$1\times10^{-5}$ & 1.90 & 0.95 \\
$2\times10^{-5}$ & 19.63 & 0.00 \\
\bottomrule
\end{tabular}}
\caption{The 8B rank-to-behavior map across the learning-rate sweep. No rate drops rank and argmax agreement together: at $8\times10^{-6}$ and $1\times10^{-5}$ the rank collapses while agreement stays high, and at $2\times10^{-5}$ agreement collapses while the rank is preserved, so the coupling seen at 4B does not transfer.}
\label{tab:decouple}
\end{table}

\section{The Positive Control in Detail}
\label{app:pc}

The positive control keeps the harness, reward, and recipe of the 4B text grid and changes only the tasks. We profiled the supervised policy on all 119 MiniWoB tasks and selected ten with a positive sampled-minus-greedy gap (Table~\ref{tab:headroom}). The same GRPO recipe, run over a small learning-rate sweep at two seeds, climbs on this set: five of six configurations rise and the strongest gains a credible 22 points (Table~\ref{tab:pcconfig}). Reading the late-layer rank on three checkpoints from this family shows rank health accompanying the gain and rank collapse accompanying the loss (Table~\ref{tab:rankhealth}), the constructive complement to the degrade lesion of the mastered grid.

\begin{table}[t]
\centering
\small
\setlength{\tabcolsep}{6pt}
\fit{\begin{tabular}{@{}l c c c@{}}
\toprule
Task & greedy SR & sampled SR & $\Delta_{\mathrm{head}}$ \\
\midrule
draw-line & 0.00 & 0.58 & $+0.58$ \\
count-sides & 0.00 & 0.33 & $+0.33$ \\
click-menu & 0.20 & 0.50 & $+0.30$ \\
use-autocomplete & 0.20 & 0.46 & $+0.26$ \\
\bottomrule
\end{tabular}}
\caption{Example headroom tasks. The reward is reachable by sampling that greedy decoding misses, so $\Delta_{\mathrm{head}}=\mathrm{SR}_{\mathrm{sample}}-\mathrm{SR}_{\mathrm{greedy}}>0$, the criterion the ten selected tasks satisfy and the mastered grid does not.}
\label{tab:headroom}
\end{table}

\begin{table}[t]
\centering
\small
\setlength{\tabcolsep}{5pt}
\fit{\begin{tabular}{@{}l c c c@{}}
\toprule
Configuration & SR & $\Delta$SR & McNemar $p$ \\
\midrule
Supervised baseline & 20.0 & --- & --- \\
lr $1\times10^{-5}$, seed 1 & 14.0 & $-6.0$ & --- \\
lr $3\times10^{-6}$, seed 1 & 32.0 & $+12.0$ & 0.070 \\
lr $5\times10^{-6}$, seed 1 & 34.0 & $+14.0$ & 0.016 \\
lr $1\times10^{-5}$, seed 2 & 38.0 & $+18.0$ & 0.012 \\
lr $3\times10^{-6}$, seed 2 & 38.0 & $+18.0$ & 0.004 \\
lr $5\times10^{-6}$, seed 2 & 42.0 & $+22.0\ [+8,+40]$ & 0.007 \\
\bottomrule
\end{tabular}}
\caption{Per-configuration positive control on the ten headroom tasks (50 matched episodes). Five of six configurations climb, four credibly, and the low rate that is a no-op on the mastered grid gains $12$ to $18$ points here.}
\label{tab:pcconfig}
\end{table}

\begin{table}[t]
\centering
\small
\setlength{\tabcolsep}{6pt}
\fit{\begin{tabular}{@{}l c c c@{}}
\toprule
Checkpoint & SR & L35 rank & argmax agr \\
\midrule
Supervised start & 20.0 & 6.88 & 1.00 \\
Climb (lr $5\times10^{-6}$, s2) & 42.0 & 7.14 & 0.96 \\
Degrade (lr $1\times10^{-5}$, s1) & 12.0 & 3.21 & 0.56 \\
\bottomrule
\end{tabular}}
\caption{Rank health on the positive-control family. The climbing checkpoint keeps its late-layer rank at the supervised level while success rises, and the degrading checkpoint's rank collapses as success falls, so rank tracks capability in both directions at 4B. These are single checkpoints and the upward move is small.}
\label{tab:rankhealth}
\end{table}

\section{Figure Gallery}
\label{app:figures}

This section collects the supporting figures that do not appear in the main text, grouped by the role they play in the argument. All success rates are clean greedy decoding on the 11-task text track (11 tasks by 5 seeds, 55 episodes per run) unless a caption states otherwise, and all intervals are Wilson $95\%$ or task-clustered paired bootstrap intervals as noted.

\subsection{Controlled null}

\begin{figure}[t]
  \centering
  \includegraphics[width=\linewidth]{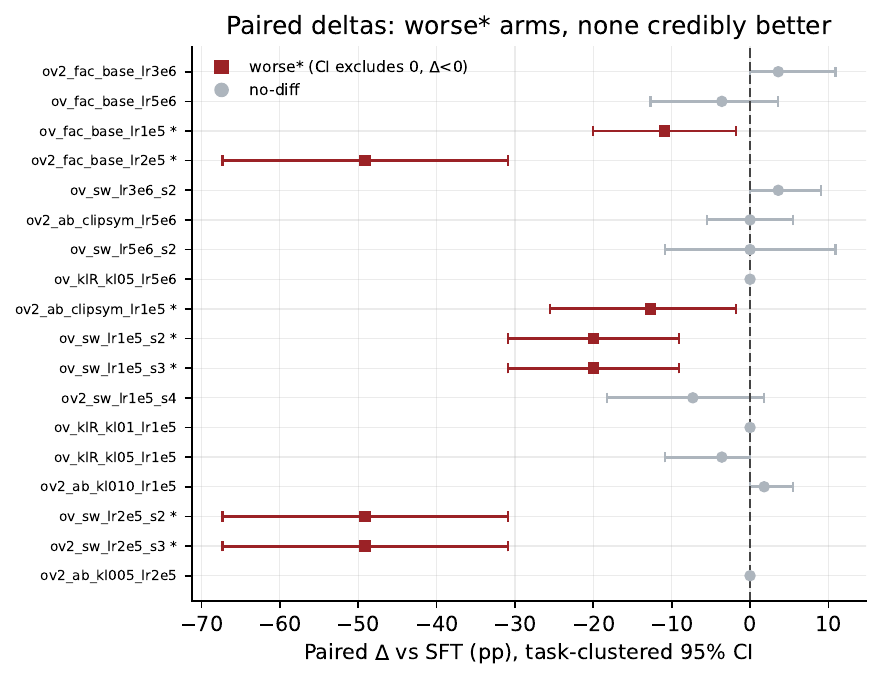}
  \caption{Paired re-analysis of every grid arm against SFT on the 55 matched (task, seed) episodes, using a task-clustered bootstrap $95\%$ interval and an exact McNemar test on the discordant pairs. No arm is credibly better than SFT. The nominal winners at lr $3\times10^{-6}$ ($+3.6$pp) have intervals $[+0.0,+10.9]$ and $[+0.0,+9.1]$ with McNemar $p=0.688$ and $0.500$. The moving regimes are credibly worse: pooled lr $1\times10^{-5}=-15.0$pp $[-20.9,-9.5]$ and pooled lr $2\times10^{-5}=-49.1$pp $[-62.7,-36.4]$. The KL-rescued arms have discordant counts $b/c=0/0$, i.e. per-episode outcomes identical to SFT on all 55 episodes.}
  \label{fig:paired_forest}
\end{figure}

\subsection{Robustness}

\begin{figure}[t]
  \centering
  \includegraphics[width=\linewidth]{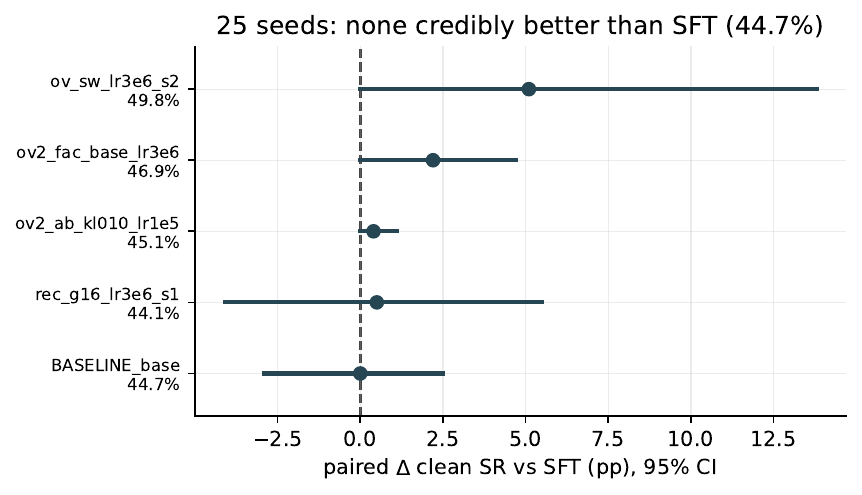}
  \caption{The null under a five-fold increase in evaluation seeds (25 seeds, 275 episodes per config). The 25-seed SFT baseline is $44.7\%$ $[39.0,50.6]$ (123/275), slightly below the 5-seed $49.1\%$. Under the task-clustered paired bootstrap no config is credibly better: the largest nominal gap is $+5.1$pp with McNemar $p=0.001$ but a bootstrap interval whose lower bound touches $0$, so it is not credible and is a single training seed.}
  \label{fig:moreseeds_forest}
\end{figure}

\begin{figure}[t]
  \centering
  \includegraphics[width=\linewidth]{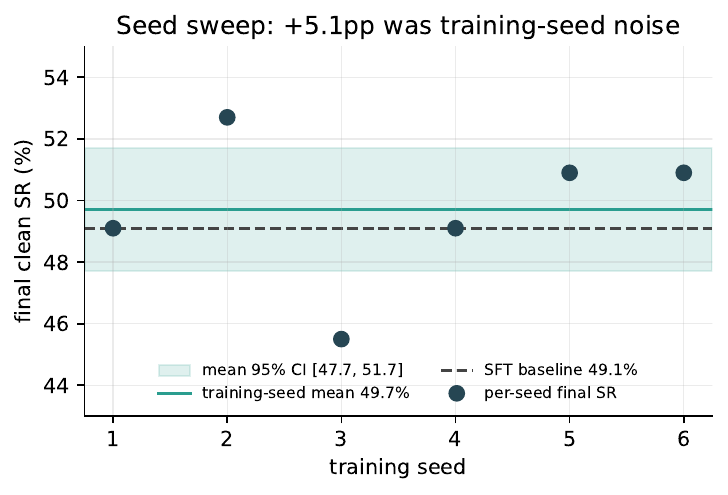}
  \caption{Training-seed replication of the one borderline arm ($+5.1$pp at 25 eval seeds). The exact configuration (SFT-init, constant lr $3\times10^{-6}$, $G=8$, 15 rounds) was retrained across six training seeds and clean-evaluated at the final checkpoint: $49.1, 52.7, 45.5, 49.1, 50.9, 50.9\%$. The mean is $49.7\%\pm2.5$ (sd), $95\%$ CI $[47.7,51.7]$, sitting on the SFT baseline of $49.1\%$, so the $+5.1$pp was training-seed noise.}
  \label{fig:seedsweep}
\end{figure}

\begin{figure}[t]
  \centering
  \includegraphics[width=\linewidth]{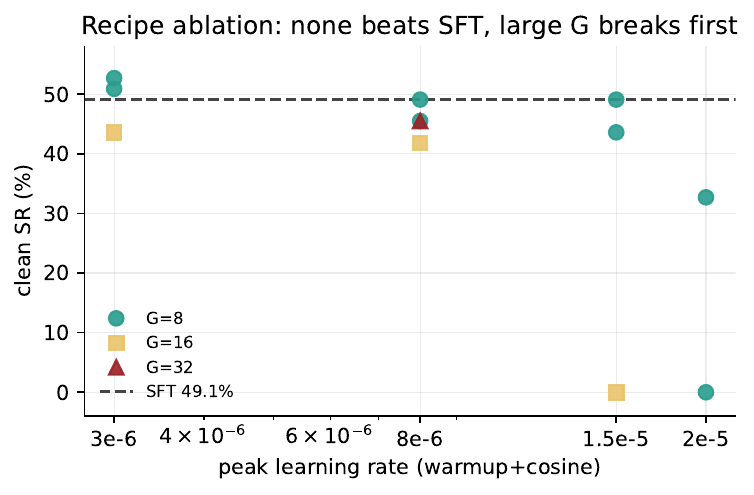}
  \caption{Recipe ablation adding a warmup ($15\%$) plus cosine schedule and sweeping group size $G\in\{8,16,32\}$, paired against SFT ($49.1\%$). No run credibly beats SFT. The schedule removes the collapse pathology at moderate learning rate (constant-lr $1\times10^{-5}$ degraded and $2\times10^{-5}$ collapsed in the grid; with the schedule, $8\times10^{-6}$ and $1.5\times10^{-5}$ at $G=8$ become no-diff), but recovery is not learning. Larger $G$ does not help and makes high-lr collapse worse.}
  \label{fig:recipe}
\end{figure}

\begin{figure}[t]
  \centering
  \includegraphics[width=\linewidth]{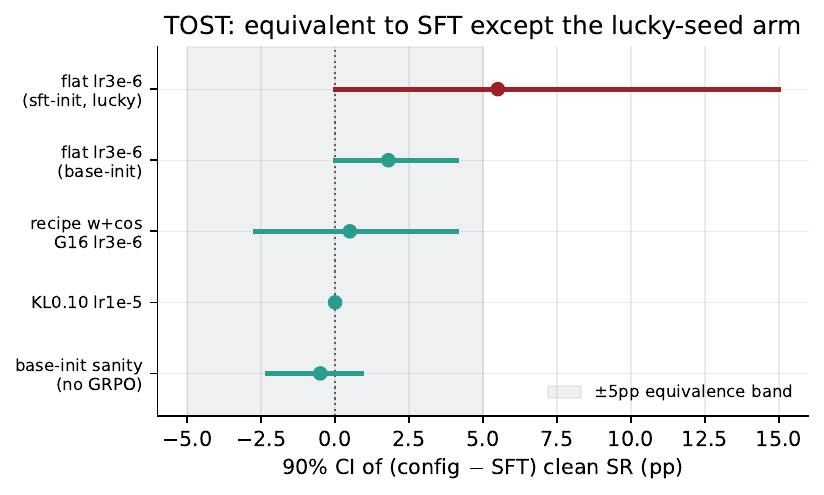}
  \caption{Two one-sided equivalence at the 25-seed grain: a config is declared equivalent to SFT within $\pm\delta$ when its task-clustered paired $90\%$ interval of (config minus SFT) lies inside $[-\delta,+\delta]$. Four of five top configs are equivalent within $\pm5$pp: base-init flat $[+0.0,+4.1]$; KL0.10 $[+0.0,+0.0]$ (episode-identical); recipe $G16$ $[-2.7,+4.1]$; base-init sanity $[-2.3,+0.9]$. The lone exception is the lucky-seed arm, $[+0.0,+15.0]$, the same seed noise resolved by the six-seed replication (Fig.~\ref{fig:seedsweep}). This is a positive equivalence claim, not merely a failure to reject.}
  \label{fig:tost}
\end{figure}

\begin{figure}[t]
  \centering
  \includegraphics[width=\linewidth]{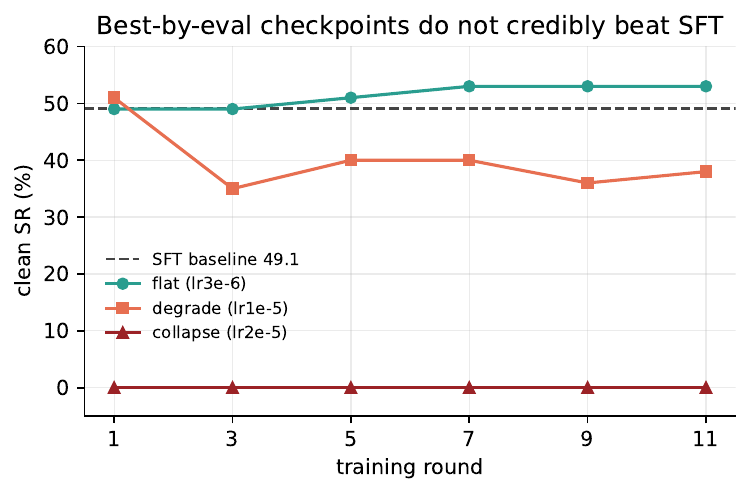}
  \caption{Best-by-eval checkpoint trajectory for the three regime exemplars, clean-evaluated at every saved round. Best-by-eval selection surfaces nominal nudges (the flat run drifts to a max of $52.7\%$ by rounds 7 to 11; the degrade run reads $50.9\%$ at round 1 before falling to $35$--$40\%$), but none survives the episode-matched paired test: flat max $+3.6$pp $[+0.0,+9.1]$ ($p=0.500$), degrade r1 $+1.8$pp $[-9.1,+12.7]$ ($p=1.000$), collapse $-49.1$pp. The null is not an artifact of scoring an already-broken final checkpoint.}
  \label{fig:bestckpt}
\end{figure}

\subsection{Mechanism}

\begin{figure}[t]
  \centering
  \includegraphics[width=\linewidth]{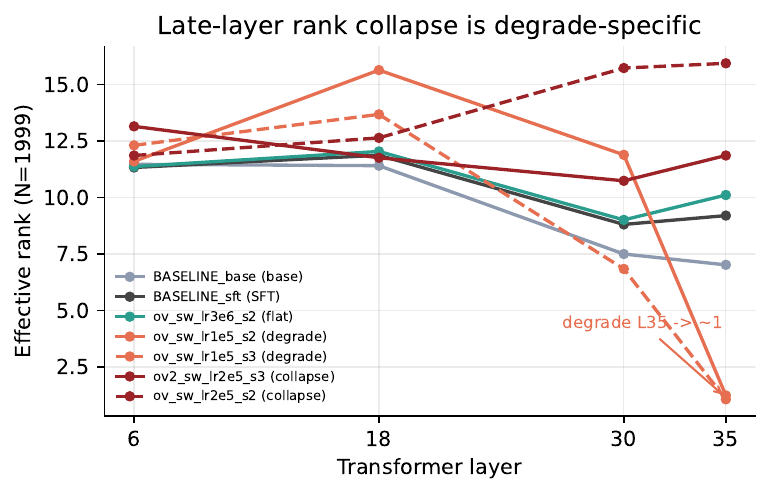}
  \caption{Fixed-state effective rank by network depth (layers 6, 12, 18, 24, 30, 35) on identical inputs, localizing the degrade lesion to the late layers. Baselines and flat runs hold a late-layer (L35) effective rank of about $6.4$ to $6.7$, whereas the no-KL degrade runs collapse L35 to about $1.0$ to $1.2$ while earlier layers (L6 to L18) stay healthy. Collapse runs preserve L35 rank, the opposite lesion. The dissociation survives at $N=1999$ states, so it is not a small-$N$ artifact.}
  \label{fig:rank_layers}
\end{figure}

\begin{figure}[t]
  \centering
  \includegraphics[width=\linewidth]{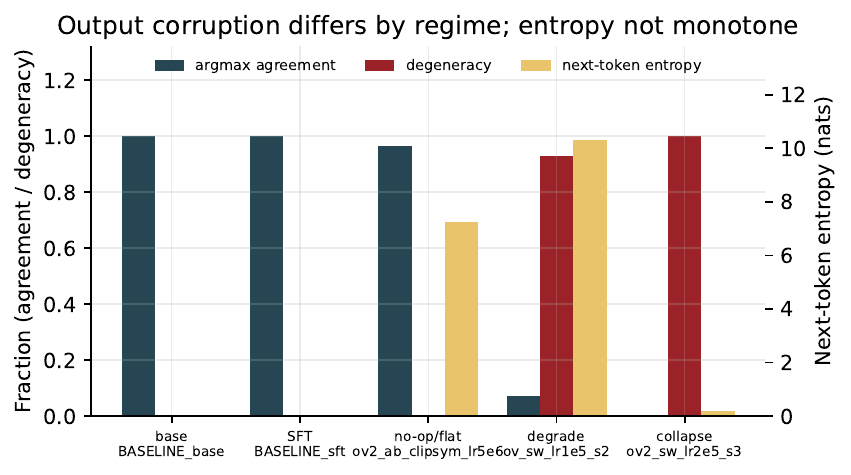}
  \caption{Fixed-state output behavior on identical decision states: baseline argmax agreement, degeneracy (unparseable or no-op fraction), and next-token entropy. Baselines and flat runs preserve the policy (agreement about $0.95$ to $1.0$, degeneracy near $0$) even when entropy is inflated, so entropy inflation alone is harmless under greedy decoding. Degrade corrupts the argmax (agreement $0.07$ to $0.5$, degeneracy $0.5$ to $0.93$); collapse destroys it entirely (agreement $0$, degeneracy $0.8$ to $1.0$) in two flavors, high-entropy garble and low-entropy mode collapse.}
  \label{fig:output_behavior}
\end{figure}

\begin{figure}[t]
  \centering
  \includegraphics[width=\linewidth]{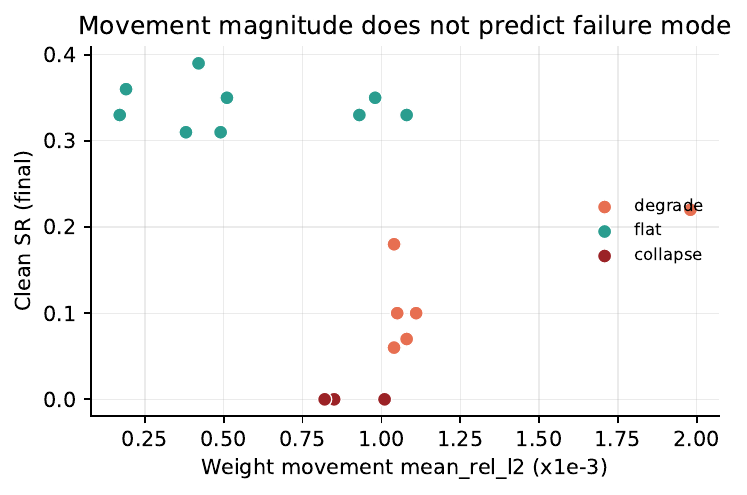}
  \caption{Weight-movement magnitude relative to initialization (mean relative $L_2$ and sparse fraction) versus learning rate and regime, showing that magnitude does not predict the failure mode. Low-lr flat runs move least. Collapse moves less in $L_2$ than degrade yet is far more destructive. The single largest total movement in the grid belongs to the KL-rescued run, which is behaviorally at the SFT baseline ($49.1\%$). The outcome is not monotone in how far the weights move; direction and structure decide it.}
  \label{fig:movement}
\end{figure}

\begin{figure}[t]
  \centering
  \includegraphics[width=\linewidth]{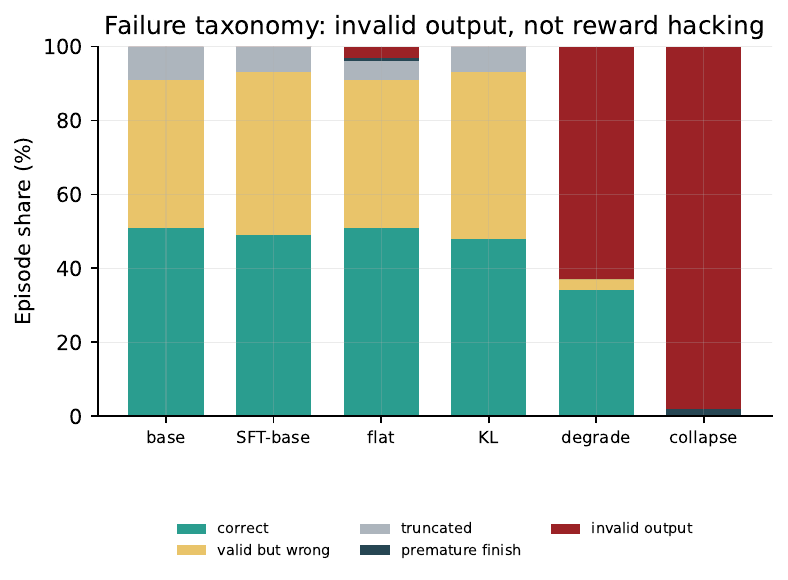}
  \caption{Per-episode failure-mode taxonomy by regime (correct, valid-but-wrong target, premature-finish, invalid or unparseable, truncated). Reward-hacking by premature finishing stays at roughly $0$ to $2\%$ across every regime, so the null is not explained by reward-gaming. The moving regimes fail through output degeneration: degrade is $63\%$ invalid and collapse is $98\%$ invalid. Flat and KL runs preserve the SFT structure (about $50\%$ correct and $40\%$ valid-but-wrong), the base capability ceiling rather than a new failure.}
  \label{fig:taxonomy}
\end{figure}

\subsection{Generalization}

\begin{figure}[t]
  \centering
  \includegraphics[width=\linewidth]{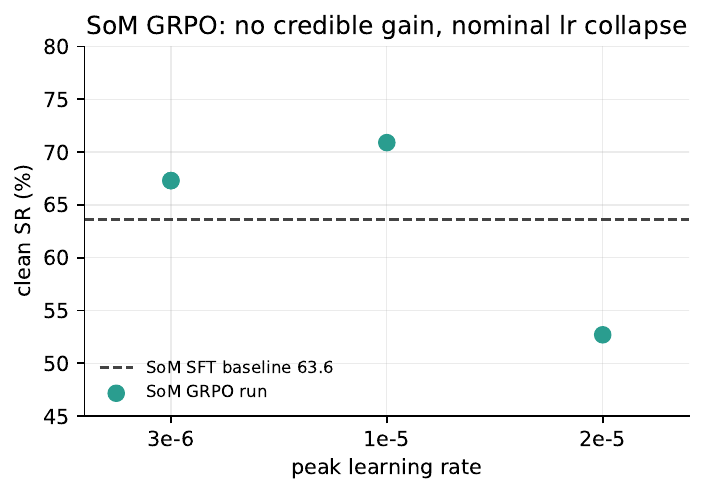}
  \caption{Set-of-Marks (vision) learning-rate sweep for the 4B model, paired against the SoM SFT baseline of $63.6\%$ $[50.4,75.1]$ (35/55), higher than the text baseline and itself a text-versus-SoM asymmetry. No SoM config credibly beats the baseline: lr $3\times10^{-6}$ is $+3.6$pp ($p=0.500$), lr $1\times10^{-5}$ is a nominal $+7.3$pp ($p=0.219$), and lr $2\times10^{-5}$ is $-10.9$pp ($p=0.070$, interval touching $0$), a nominal-but-not-credible degrade. The no-gain null generalizes to SoM; the credible lr-collapse established on text is only nominal here.}
  \label{fig:som_lr}
\end{figure}

\begin{figure}[t]
  \centering
  \includegraphics[width=\linewidth]{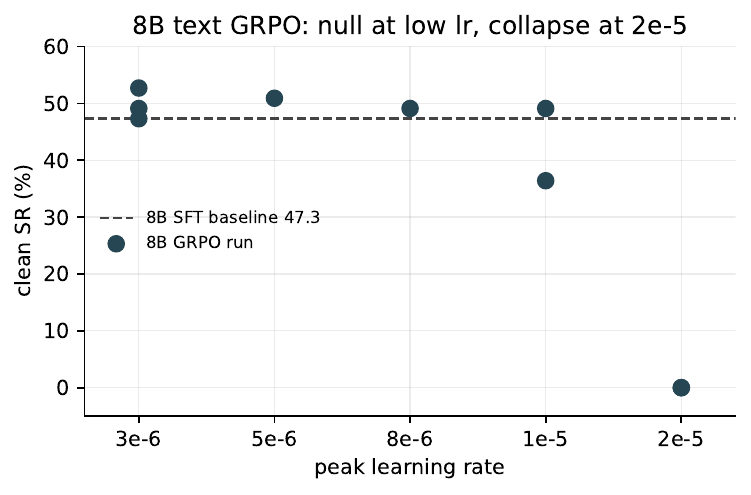}
  \caption{Text-track learning-rate sweep for the 8B model, paired against the 8B-SFT baseline of $47.3\%$ $[34.7,60.2]$ (26/55). No 8B config credibly beats the baseline. The low-lr arms give the familiar nominal nudge (lr $3\times10^{-6}$ up to $52.7\%$, not credible), while the moving regimes break: lr $1\times10^{-5}$ seed 2 is credibly worse ($36.4\%$, $p=0.031$) and both lr $2\times10^{-5}$ runs collapse to $0.0\%$ ($p<0.001$). The null and the high-lr breakage carry over from 4B to 8B.}
  \label{fig:8b_lr}
\end{figure}

\begin{figure}[t]
  \centering
  \includegraphics[width=\linewidth]{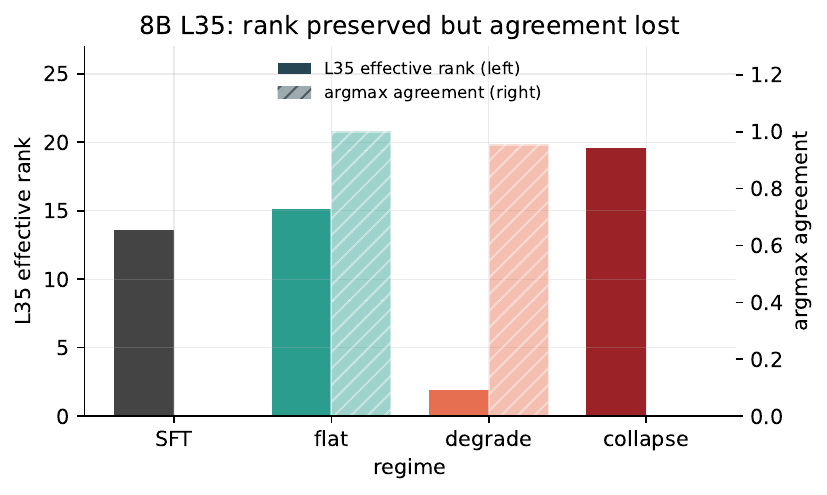}
  \caption{8B mechanism interpretability on $N=372$ fixed states. The late-layer (L35) effective rank matches the 4B pattern: SFT $13.6$, flat $15.1$, degrade collapsed to $1.9$, collapse preserved at $19.6$. The honest scale nuance is that the 8B degrade keeps argmax agreement high ($0.954$) with intact success despite the rank collapse, whereas the 4B degrade drops agreement to $0.07$ to $0.5$. So the rank leg of the dissociation reproduces at 8B while the rank-to-behavior coupling does not.}
  \label{fig:8b_interp_bars}
\end{figure}

\begin{figure}[t]
  \centering
  \includegraphics[width=\linewidth]{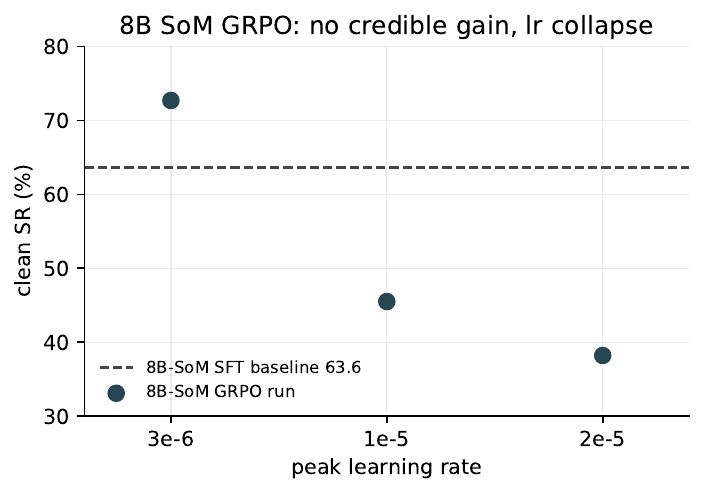}
  \caption{8B Set-of-Marks learning-rate sweep, paired against the 8B-SoM-SFT baseline of $63.6\%$ $[50.4,75.1]$ (35/55). No 8B-SoM config credibly beats the baseline: lr $3\times10^{-6}$ is the same low-lr nominal nudge ($+9.1$pp, $p=0.062$, single seed), while lr $1\times10^{-5}$ ($-18.2$pp, $p=0.002$) and lr $2\times10^{-5}$ ($-25.5$pp, $p=0.001$) are both credibly worse. The lowest learning rate gives a nominal positive nudge that never clears the paired interval, here as at 4B.}
  \label{fig:8b_som}
\end{figure}

\section{Algorithms and Additional Definitions}
\label{app:algo}

This section states the three procedures the main text refers to by name, and recaps two estimators used in the evaluation. Algorithm~\ref{alg:grpo} is the update we sweep; the only knobs the grid varies are the learning rate $\eta_t$, the KL weight $\beta$, the clip pair $(\epsilon_{\mathrm{lo}},\epsilon_{\mathrm{hi}})$, the seed, and the initialization of $\theta$. Algorithm~\ref{alg:bootstrap} is the interval behind every ``credible'' verdict, and Algorithm~\ref{alg:graft} is the causal test behind the localization claim.

\begin{algorithm}[t]
\caption{GRPO update for the web agent}
\label{alg:grpo}
\begin{algorithmic}[1]
\Require policy $\pi_\theta$, frozen initialization $\pi_{\mathrm{ref}}$, prompts, group size $G$, clip $(\epsilon_{\mathrm{lo}},\epsilon_{\mathrm{hi}})$, KL weight $\beta$, schedule $\eta_t$
\For{each round $t$ and prompt $x$}
  \State sample $G$ rollouts $\{\tau_i\}$ from $\pi_{\theta_{\mathrm{old}}}(\cdot\mid x)$
  \State $r_i \gets \mathbf{1}[\operatorname{success}(\tau_i)]$ \Comment{sparse terminal reward}
  \State $A_i \gets r_i - \tfrac{1}{G}\sum_{j} r_j$ \Comment{group mean-centered, no std}
  \For{each completion token $c$ of each rollout}
    \State $\rho_c \gets \pi_\theta(y_c\mid x,y_{<c}) / \pi_{\theta_{\mathrm{old}}}(y_c\mid x,y_{<c})$
    \State $g_c \gets \min\!\big(\rho_c A,\ \operatorname{clip}(\rho_c,1-\epsilon_{\mathrm{lo}},1+\epsilon_{\mathrm{hi}})\,A\big)$
  \EndFor
  \State $\mathcal{L} \gets -\tfrac{1}{C}\sum_c g_c + \beta\,\widehat{D}_{\mathrm{KL}}(\pi_\theta \,\|\, \pi_{\mathrm{ref}})$
  \State $\theta \gets \theta - \eta_t\,\nabla_\theta \mathcal{L}$
\EndFor
\end{algorithmic}
\end{algorithm}

\begin{algorithm}[t]
\caption{Task-clustered paired bootstrap for $\Delta\mathrm{SR}$}
\label{alg:bootstrap}
\begin{algorithmic}[1]
\Require matched per-episode outcomes for an arm and for SFT over tasks $\mathcal{T}$ ($|\mathcal{T}|=11$) and seeds, resamples $B$
\For{$b=1$ to $B$}
  \State draw $|\mathcal{T}|$ task clusters from $\mathcal{T}$ with replacement
  \State $\Delta_b \gets \mathrm{SR}_{\mathrm{arm}} - \mathrm{SR}_{\mathrm{SFT}}$ over the drawn clusters
\EndFor
\State \Return the $2.5$ and $97.5$ percentiles of $\{\Delta_b\}$
\Statex \textbf{Verdict:} \emph{credibly better/worse} iff the interval excludes $0$; \emph{equivalent within} $\delta$ iff the $90\%$ interval lies inside $[-\delta,+\delta]$
\end{algorithmic}
\end{algorithm}

\begin{algorithm}[t]
\caption{Causal component grafting}
\label{alg:graft}
\begin{algorithmic}[1]
\Require trained weights $\theta$, initialization $\theta^{\mathrm{init}}$, component group $\mathcal{G}$ (attention, MLP, embedding, or a random parameter set of matched size)
\State $\theta' \gets \theta$
\State $\theta'_{\mathcal{G}} \gets \theta^{\mathrm{init}}_{\mathcal{G}}$ \Comment{restore one group, keep the rest}
\State \Return $\mathrm{SR}(\pi_{\theta'})$ \Comment{recovery attributable to $\mathcal{G}$}
\end{algorithmic}
\end{algorithm}

Two estimators complete the evaluation. The headroom criterion that selects the positive-control tasks compares the greedy success to the mean success under temperature sampling of the supervised policy,
\begin{equation}
\Delta_{\mathrm{head}} = \tfrac{1}{K}\textstyle\sum_{k} \mathbf{1}[\operatorname{success}(\tau_k)] - \mathbf{1}[\operatorname{success}(\tau_{\mathrm{greedy}})],
\end{equation}
averaged over evaluation seeds, and a task is climbable when $\Delta_{\mathrm{head}}>0$. The random-component null for grafting draws a set $\mathcal{R}$ of non-attention parameters with $|\mathcal{R}|$ equal to the attention parameter count, restores $\mathcal{R}$ from initialization, and repeats the draw $24$ times to form the null distribution whose mean and 95th percentile Table~\ref{tab:graft} reports; a real graft is a specific cause only when its recovery exceeds that percentile.

\end{document}